\documentclass[sigconf]{acmart}

\usepackage{amsfonts}
\usepackage{amsmath}
\usepackage{amssymb}
\usepackage{amsthm}
\usepackage{algorithm2e}
\usepackage{algorithmic}
\usepackage{bm}
\usepackage{caption}
\usepackage{courier}
\usepackage{dsfont}
\usepackage{enumitem}
\usepackage{float}
\usepackage{graphicx}
\usepackage{helvet}
\usepackage{hyperref}
\usepackage{multirow}
\usepackage{siunitx}
\usepackage{soul}
\usepackage{subfigure}
\usepackage{textcomp}
\usepackage[normalem]{ulem}
\usepackage{wrapfig}
\usepackage{xcolor}
\usepackage{url}
\usepackage{xspace}

\usepackage[disable]{todonotes}

\newcommand{\ap}{\todo[author=AP, color=yellow,inline]}

\hypersetup{
    colorlinks=true,
    linkcolor=blue,
    filecolor=magenta,      
    urlcolor=cyan,
}
\makeatletter
\DeclareRobustCommand\bmvaOneDot{\futurelet\@let@token\bmv@onedotaux}
\def\bmv@onedotaux{\ifx\@let@token.\else.\null\fi\xspace}
\makeatother
\def\eg{\emph{e.g}\bmvaOneDot}

\def\etal{\emph{et al}\bmvaOneDot}
\def\etc{\emph{etc}\bmvaOneDot}
\def\ie{\emph{ie}\bmvaOneDot}
\newcommand{\argmin}[1]{\underset{#1}{\operatorname{argmin}}\;}

\AtBeginDocument{%
  \providecommand\BibTeX{{%
    \normalfont B\kern-0.5em{\scshape i\kern-0.25em b}\kern-0.8em\TeX}}}

\setcopyright{acmcopyright}
\copyrightyear{2019} 
\acmYear{2019} 
\acmConference[BuildSys '19]{The 6th ACM International Conference on Systems for Energy-Efficient Buildings, Cities, and Transportation}{November 13--14, 2019}{New York, NY, USA}
\acmBooktitle{The 6th ACM International Conference on Systems for Energy-Efficient Buildings, Cities, and Transportation (BuildSys '19), November 13--14, 2019, New York, NY, USA}
\acmDOI{10.1145/3360322.3360853}
\acmISBN{978-1-4503-7005-9/19/11}

\acmPrice{}

\begin{document}

\title{COLTRANE: ConvolutiOnaL TRAjectory NEtwork for Deep Map Inference}



\author{Arian Prabowo}
\affiliation{%
    \institution{RMIT University, DATA61/CSIRO}
    \city{Melbourne}
    \country{Australia}
}
\email{arian.prabowo@rmit.edu.au}
\orcid{}

\author{Piotr Koniusz}
\affiliation{%
    \institution{DATA61/CSIRO, ANU}
    \city{Canberra}
    \country{Australia}
}
\email{piotr.koniusz@data61.csiro.au}

\author{Wei Shao}
\affiliation{%
    \institution{RMIT University}
    \city{Melbourne}
    \country{Australia}
}
\email{wei.shao@rmit.edu.au}

\author{Flora D. Salim}
\affiliation{%
    \institution{RMIT University}
    \city{Melbourne}
    \country{Australia}
}
\email{flora.salim@rmit.edu.au}

\renewcommand{\shortauthors}{}

\begin{abstract}
%
The process of automatic generation of a road map from GPS trajectories,  called map inference, remains a challenging task to perform on a  geospatial data from a variety of domains as the majority of existing studies  focus on road maps in cities. 
Inherently, existing algorithms are not  guaranteed to work on unusual geospatial sites, such as  an airport tarmac, pedestrianized paths and shortcuts, or animal migration routes, \etc.
Moreover, deep learning has not been explored well enough for such tasks.  

This paper introduces COLTRANE,  ConvolutiOnaL TRAjectory NEtwork, a novel deep map inference framework
which operates on GPS trajectories collected in various environments.
This framework includes an Iterated Trajectory Mean Shift (ITMS) module to localize road centerlines, which copes with noisy GPS data points.
Convolutional Neural Network trained on our novel trajectory descriptor is then
introduced into our framework to detect and accurately classify junctions for refinement of the road maps.
COLTRANE yields up to 37\% improvement in F1 scores over existing methods on two distinct real-world datasets: city roads and airport tarmac.

\end{abstract}





\maketitle

\section{Introduction}

Traditionally, building a digital map includes digitizing current paper maps, having surveyors to visit the grounds and manually edit the map, and using the aerial photography \cite{worrall2007automated}. However, the combination of traditional methods can be costly and the physical access to sites may be restricted. Roads often suffer from congestion and downtime due to maintenance or ongoing constructions, making frequent updates by classical approaches costly.

\sloppy{Recently, the number of datasets which consist of GPS datapoints has been growing \cite{Shao2018,cruz2015grouping,qin2019solving}. Availability of such a data presents researchers with an opportunity to design new algorithms for map inference from the GPS data. Approach \cite{guo2007towards} calls this process as `data recycling' while \cite{biagioni2012inferring} calls it as map inference. With algorithms for high quality map inference, obtaining accurate digital maps and their updates becomes viable and cost-effective.}

The process of map inference poses a number of challenges. The GPS signal is noisy, especially in urban areas \cite{biagioni2012map}, making extraction of road segments, and the detection  of junctions, a non-trivial pursuit. The datasets are often unbalanced as some roads are travelled frequently while other roads (\eg, rural) are not. Thus, in some cases, it may be hard to determine if a collection of datapoints represent spurious noises or sparsely travelled routes.

Further, the map inference is often addressed with a technique specific to a given site or only tested with geospatial data from a particular domain \eg, GPS trajectories from road vehicles or predefined route networks. Therefore, standard map-matching techniques could be used when the movement data comes from popular well mapped urban areas. However, the map inference becomes difficult when the geospatial trajectory data comes from commercial non-public areas or precincts, such as from airport tarmac areas \cite{shao2019OnlineAirTrajClus} and parking spaces \cite{shao2016clustering}. Often, in these cases, a reliable up-to-date map is non available and existing map inference methods fail. Below, we familiarize the reader with existing approaches.


A seminal paper on map inference \cite{edelkamp2003route} uses a modified k-means algorithm to estimate road centerlines. Others followed and improved upon their baseline \cite{schroedl2004mining,worrall2007automated,agamennoni2011robust}. Recently, approach \cite{chen2016city} employed a modified mean shift instead of k-means.

Subsequently, many computer vision based approaches, first pioneered by \cite{davies2006scalable} and then followed by \cite{chen2008roads,shi2009automatic,jang2010map}, convert GPS trajectories to an image, compute a 2D histogram, and use a variety of different image processing tools for post-processing. Following the above direction, approach \cite{biagioni2012map} combined aspects of existing algorithms and introduced so-called grey-scale skeletonization that models uncertainty of the road centerlines as gray-scale image representation (the state of the art until recently). Subsequently, approach by Chen \etal \cite{chen2016city} integrated the prior knowledge of roads into the inference step. Moreover,  Chen \etal \cite{chen2016city} used of a modified version of a popular local image descriptors SIFT \cite{lowe1999object}, called Traj-SIFT, which works directly on a directed graph data. 
%
%
However, their algorithm has not been applied to datasets from commercial non-public precincts such as airports. We are not aware of any map inference approaches  applied to areas which lack a well-established road-network.

Moreover, in the decade of AI celebrating Convolutional Neural Networks (CNN) \cite{krizhevsky2012imagenet}, it appears CNNs have not yet been used for the road map inference despite of their learning ability.  In this paper, we introduce COLTRANE, ConvolutiOnaL TRAjectory NEtwork, a novel deep learning framework for the map inference from GPS trajectory, which produces an annotated directed graph.

In COLTRANE, 
we improve upon an existing variant of mean shift by appropriating it for complex trajectory data. We call it Iterated Trajectory Mean Shift Sampling (ITMS), and we use it to approximate the road centerlines by generating centerline points which are then connected to form a road map, represented by a directed graph.  As we treat the centerline points as nodes of the graph, each node may constitute on a different kind of road junction (or segment). Thus, the number of road lanes coming in/out of it correlates with the node degree in the graph. We apply a CNN to predict the degree of each node to infer the road connectivity and we classify each node into a junction type (several kinds) or a straight road segment. We develop novel trajectory descriptors as the input to the CNN. We evaluate our method on two distinct real-world city and airport datasets. 

In what follows, we detail our contributions, discuss related works and our notations. Next, we discuss challenges of map inference as well as the uniqueness of the airport dataset. Then we describe our framework COLTRANE. Lastly, we present our results on two datasets, using visual and quantitative evaluations.

\subsection{Contributions}


We  propose COLTRANE, a deep learning framework for map inference from trajectories which is generic and adaptable to both city and airport tarmac environments. COLTRANE contains proposed by us three components:

\renewcommand{\labelenumi}{\roman{enumi}.}
\begin{enumerate}
{
%
%
\item Iterated  Traj-Mean Shift (ITMS) algorithm which  incorporates the orientation of the motion of GPS points during their clustering process.
\item Trajectory descriptors a.k.a. features or  feature maps, which contain counts of GPS coordinates, average of $x$- and $y$-directional velocities. 
\item CNN applied by us for the first time to trajectory maps for the purpose of the junction type classification and inference of the degree of centerline points (roads going in/out of a junction) to aid the process of merging road segments.
}
\end{enumerate}


We apply COLTRANE to regular datasets collected by road vehicles, and a much more complex GPS data collected at the airport from aircraft and ground vehicles traversing the airport tarmac. 
%
%
We outperform a recent algorithm 
\cite{chen2016city} on both datasets. 
In contrast to the regular road data, airport routes are weakly defined, more noisy and poorly separated due to proximity of various lanes. Yet, we demonstrate state-of-the-art results on such a challenging data.

\section{Related Work}

Surveys on the map inference problem a.k.a. map construction, map generation, and map creation, can be found in \cite{biagioni2012inferring,ahmed2015comparison}. There is also a more recent, albeit non-technical, read \cite{gao_klaiber_patel_underwood_2019}. In addition, paper \cite{biagioni2012inferring} also introduced a directed spatial graph evaluation protocol for the purpose of evaluation of the quality of maps.

\subsection{Clustering-based approaches}
The seminal paper on map inference \cite{edelkamp2003route}, further improved by \cite{schroedl2004mining}, uses a variant of k-means to detect the road centerlines by clustering GPS datapoints that are close to each other according to the Euclidean distance and heading. A similar approach \cite{worrall2007automated} used an alternative way to infer road segments that connect the sample points. GPS was used in \cite{edelkamp2003route} with only 2D coordinates while in \cite{worrall2007automated} a new metric was introduced to find the best adjacent sample point. In \cite{agamennoni2011robust}, the altitude data was used for inference of pit mining maps. 
More complicated clustering approach  \cite{ferreira2013vector} applies clustering on vector fields rather then directly on trajectories.

Notably, Chen \etal \cite{chen2016city} developed Traj-Meanshift clustering as a better alternative to k-means, and leveraged prior knowledge regarding city road network such as the smoothness of local road segments. Similarly, \cite{huang2019road} exploited the prior knowledge regarding turning restrictions at junctions to detect them. Recently, algorithm \cite{sasaki2019road} employed yet another clustering technique called DBSCAN. 

\subsection{2D histogram based approaches}
A common family of approaches to finding the road centerlines can be summarized by approach  \cite{davies2006scalable} which 
uses a 2D histogram of interpolated GPS datapoints followed by binarization. The road centerlines were extracted by using so-called Voronoi graph. Furthermore, approach \cite{biagioni2012map} used adaptive thresholding for binarization. 

Approaches \cite{chen2008roads,shi2009automatic} used morphological operations such as the dilation and closure in place of interpolations followed by so-called skeletonization in place of the Voronoi algorithm. Furthermore, method \cite{jang2010map} clustered neighboring pixels in place of morphological operations, while approach \cite{guo2007towards} extracted the road centerlines by fitting spline curves. 
Traj-SIFT approach \cite{chen2016city} proposed modified SIFT descriptors on trajectories and employed an SVM classifier for junction detection. We note such descriptors are related to 3D human skeleton descriptors for action recognition \cite{me_tensor_eccv16,ar_domain,lei_tip_2019}.

In contrast, we use CNN on feature maps representing trajectories to infer node degrees in the road graph representing junctions.
%
\subsection{Other approaches}
Less common approaches include \cite{cao2009gps} which  simulates the physical attraction/repulsion between the GPS points to extract the centerline. More recently, \cite{stanojevic2018road,stanojevic2018robust} use similar idea and formulate the map inference as a partial graph matching. Approach \cite{he2018roadrunner} puts the focus on identifying missing road segments from the map. 
Approach \cite{niehoefer2009gps} uses the strength of mobile phone signal (beside of standard techniques) to detect bridges and tunnels. Also using the cellular network data, approach \cite{zheng2018buildings} is modelling the urban mobility in a city. 
Approach \cite{ahmed2012constructing} combined both map and partial curve matching while papers \cite{karagiorgou2012vehicle,karagiorgou2013segmentation} detect junctions and corners prior to form connections between them. Paper \cite{yang2018method} uses Delaunay triangulation and the Voronoi diagram. while approach  \cite{zheng2017topic} relies on Natural Language Processing, clustering and 2D histograms. Paper \cite{shao2019OnlineAirTrajClus}  filters out GPS points with low confidence. 
Finally, for brevity, we refer  readers to  tutorials on deep learning by Jeff Hinton \cite{hinton_cnn}.

Following  \cite{stanojevic2018road,stanojevic2018road}, we use the approach of  Chen \etal \cite{chen2016city} as the baseline to compare ourselves to due to conceptual similarities. 

To the best of our knowledge, there is no map inference paper using GPS data and deep learning. However, approach \cite{bessa2016riobusdata} performed outlier detection in bus routes by CNN using GPS data.



\section{Motivation}\label{sec:motivation}
Map inference is applicable to a variety of scenarios, verging from the traffic analysis in smart cities and rural areas to wild habitats and restricted environments \eg, constructing and updating the route networks of airport aircraft can help air traffic controllers manage aircraft landings and take-offs. It can also help airport traffic managers recognise patterns of aircraft movement to reduce the traffic congestion \cite{bertsimas1998air,kong2016urban} and detect anomalies in the routes of aircraft \cite{pusadan2017anomaly}. Moreover, it can prevent hazardous situations by ensuring ground vehicles adhere to secure routes and procedures.

Currently, companies such as Google, Apple and OpenStreetMap provide digital maps of the road networks of cities and urban areas but they are excluded from restricted areas. Those companies also spend tremendous amounts of money and human resources in manually mapping road networks into digital maps \cite{aly2014map++} which poses numerous practical issues. Airport runways and tarmac remain excluded from manual mapping, however, the Federal Aviation Administrations (FAA) have recorded the GPS trajectory data for each aircraft at United States airports, which offers an opportunity to generate  maps of aircraft ground routes \cite{chen2017short}. 

Many algorithms and frameworks have been proposed to construct the road network of cities or urban areas. None of them are applicable to the aircraft and ground vehicle trajectory data. These works can be roughly grouped into two classes:  (i) constructing the route networks of vehicles or pedestrians, where the route networks exist on the real-world map, and are represented as roads, highways or trails \cite{fu2016driving, kuntzsch2016generative} and (ii) discovering the main trajectories from the GPS data where no fixed roads, road plans and maps exist \cite{jonsen2003meta, dodge2013environmental,RN293} \eg,
animal routes or pedestrianised shortcuts. 

We note there exist practical difficulties for the map inference. The patterns of trajectory data and  evaluation metrics differ across the variety of methods for the route network construction. For existing road network reconstruction methods, route networks often match the existing road networks on the map 
while for non-existent road network such a ground truth does not exist. Researchers often group such trails into a couple of main trajectories to establish a network of popular routes. The criteria for evaluating such route networks are thus somewhat subjective. 
Compared with traditional road network map construction, inferring an airport map using aircraft GPS trajectories is more challenging. 
Firstly, airport runways are different from other sources of GPS data such as taxi or cars. The trajectories of aircraft  are more uncertain because the common roads are much narrower than airport runways, and airport ground vehicles often follow unscripted routes. Secondly, the speeds and headings of aircraft are more uncertain than those of road vehicles due to the traffic control. Thirdly, aircrafts encounter different uncontrolled situations than city vehicles. Fourthly, airport traffic changes according to  criteria such as weather, scheduling, safety \etc.  
In summary, requirements for inferring maps of  airport/city road networks   differ which necessitates our investigations. 


\subsection{Datasets}

Below, we compare the GPS noise and mobility patterns of two different real-world datasets summarized in Table \ref{tab:dataset}. We discover that the map inference on airport tarmac poses a unique set of challenges absent from the typical city map inference. 

\begin{table}[htb]
\caption{UIC and FAA datasets.}
\label{tab:dataset}
\begin{tabular}{c|cc}
\toprule
\textbf{Description (unit)} & \multicolumn{1}{c}{\textbf{UIC}} & \multicolumn{1}{c}{\textbf{FAA}} \\ \hline \hline
\multicolumn{1}{c|}{Volume (MB)}                                      & 7 & 138 \\ \hline
\multicolumn{1}{c|}{Number of points}                                & 118,364 & 1,057,688 \\ \hline
\multicolumn{1}{c|}{Number of trajectories}                           & 889 & 8,902 \\ \hline
\multicolumn{1}{c|}{Total trajectory length (\si{\hour})} & 118 & 526  \\ \hline
\multicolumn{1}{c|}{Total trajectory length (\si{\kilo\metre})} & 2,867 & 6,258 \\ \hline
\multicolumn{1}{c|}{Area coverage (\si{\square\kilo\metre})} & 9.4104 & 10.4791 \\ \hline
\multicolumn{1}{c|}{Time span} & 28 days & 24 hours \\ \hline
\multicolumn{1}{c|}{Start time}
& {\begin{tabular}[c]{@{}c@{}} 2011-04-01\\15:15:20\end{tabular}}
& {\begin{tabular}[c]{@{}c@{}} 2016-07-31\\14:00:01 \end{tabular}}  \\ \hline
\multicolumn{1}{c|}{End time}
& {\begin{tabular}[c]{@{}c@{}} 2011-04-29\\19:54:53\end{tabular}}
& {\begin{tabular}[c]{@{}c@{}} 2016-08-01\\13:59:59 \end{tabular}}  \\ \bottomrule
\end{tabular}
\end{table}

\subsubsection{UIC}


This dataset was collected by the University of Illinois at Chicago (UIC) and is available at \url{https://www.cs.uic.edu/bin/view/Bits/Software} \cite{biagioni2012map}. It was generated by GPS sensors embedded in a fleet of 13 campus shuttle buses. The mobility pattern of these buses could be grouped into 2 categories. The first and larger group consists of 11 buses that traveled regular predefined routes, thus displaying routine mobility patterns. The frequency of service varied between routes, creating a high disparity of spatial density. The other group served chartered trips with non-routine mobility. Thus, this dataset displays a highly regular patterns with only a few trajectories that deviate from predefined paths \cite{biagioni2012inferring}. 
We note that localization and mobility datasets are typically noisy. For GPS data, the sources of error include hardware, tectonic and seismic activities, seasonal cycles, and local geography \cite{nistor2016gps}. Since the buses travel between low, mid, and high rise buildings, this dataset contains a varying level of GPS noise.

\subsubsection{FAA}


This is a private dataset collected by the Federal Aviation Administration (FAA) of United State Department of Transportation, made available to us through our industry partner \cite{shao2019OnlineAirTrajClus,zhao2019intelligent}. Unlike UIC, this dataset has not been cleansed. Thus,  we add an additional data cleaning step. This dataset consists of trajectories of both airplanes and ground vehicles on the tarmac of Los Angeles Airport (LAX), the 4\textsuperscript{th} busiest airport in the world. It handled 87,534,384 passengers and 2,209,850 tonnes of cargo in year 2018 alone \cite{aci2019preliminary}. Due to the busyness of the airport, this dataset is much larger in every aspect (Table \ref{tab:dataset}), despite the fact that the data collection spans only 24 hours. The trajectories are generated with various sensors embedded in  airplanes and ground vehicles, as well as by ground sensors in the airport, such as radars. A single object can be assigned to multiple trajectories \ie, an airplane could be assigned to one trajectory during arrival, and a different one during departure due to of the change of flight number.

In LAX, most of the start and end points of trajectories are located in the runway and apron areas, the latter of which is where gates and terminals can be found. This airport has 4 runways, 2 at the north and 2 at the south \cite{noise2014los} which allow  east and west approach. In addition, there are 132 gates  spread over 9 terminals. Taxiways connect the runways, aprons, and other facilities such as hangars.

\subsection{GPS signal and noise}\label{sec:noisy_gps}

\begin{figure}[htb]
  \subfigure[An intersection from the UIC dataset (area of low-built buildings).]{
    \includegraphics[width=.4\columnwidth]{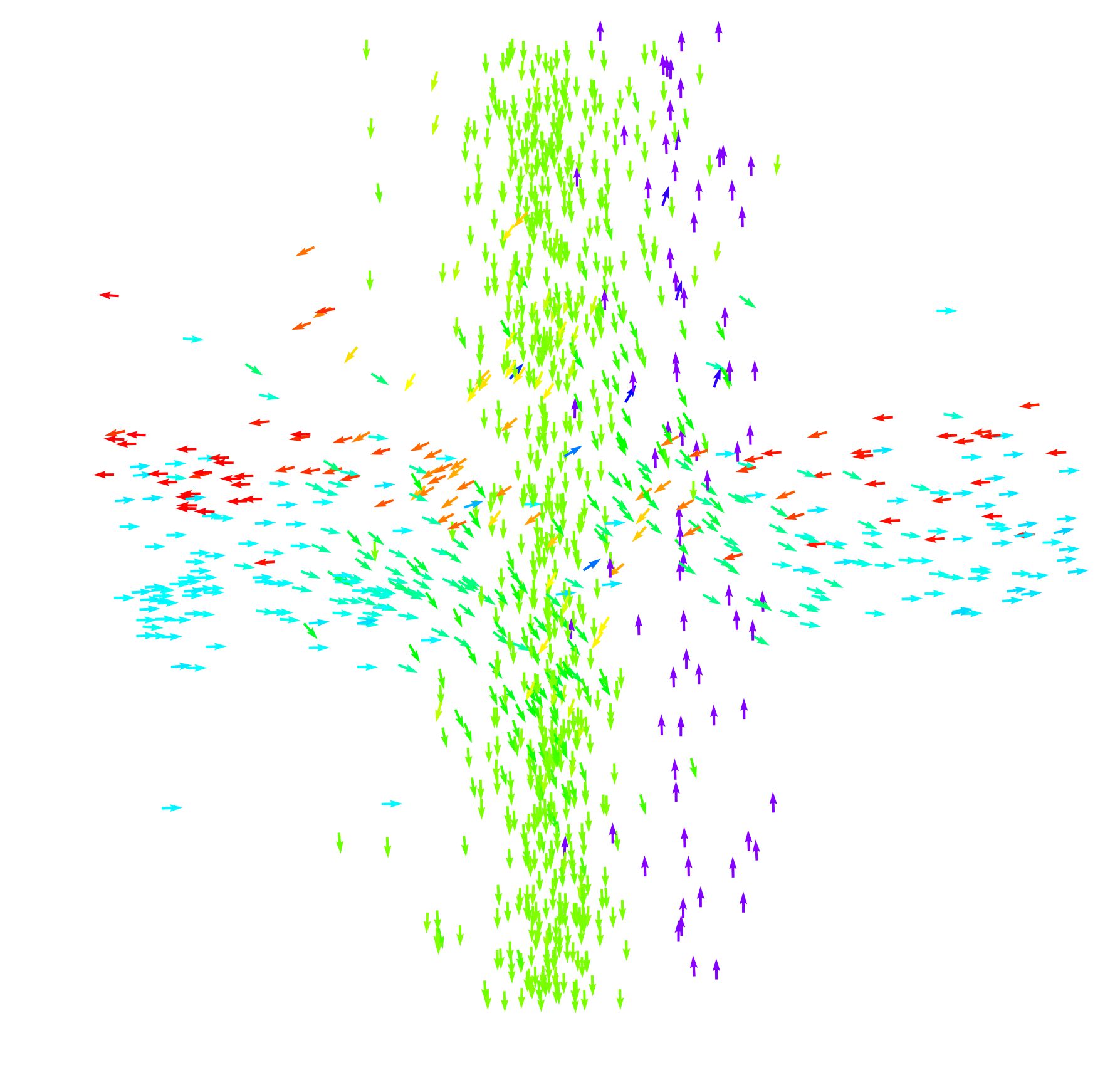}
    \label{fig:uic_lowrise}}
  \hspace{.05\columnwidth}
  \subfigure[An intersection from the UIC dataset (area of high-built buildings).]{
    \includegraphics[width=.4\columnwidth]{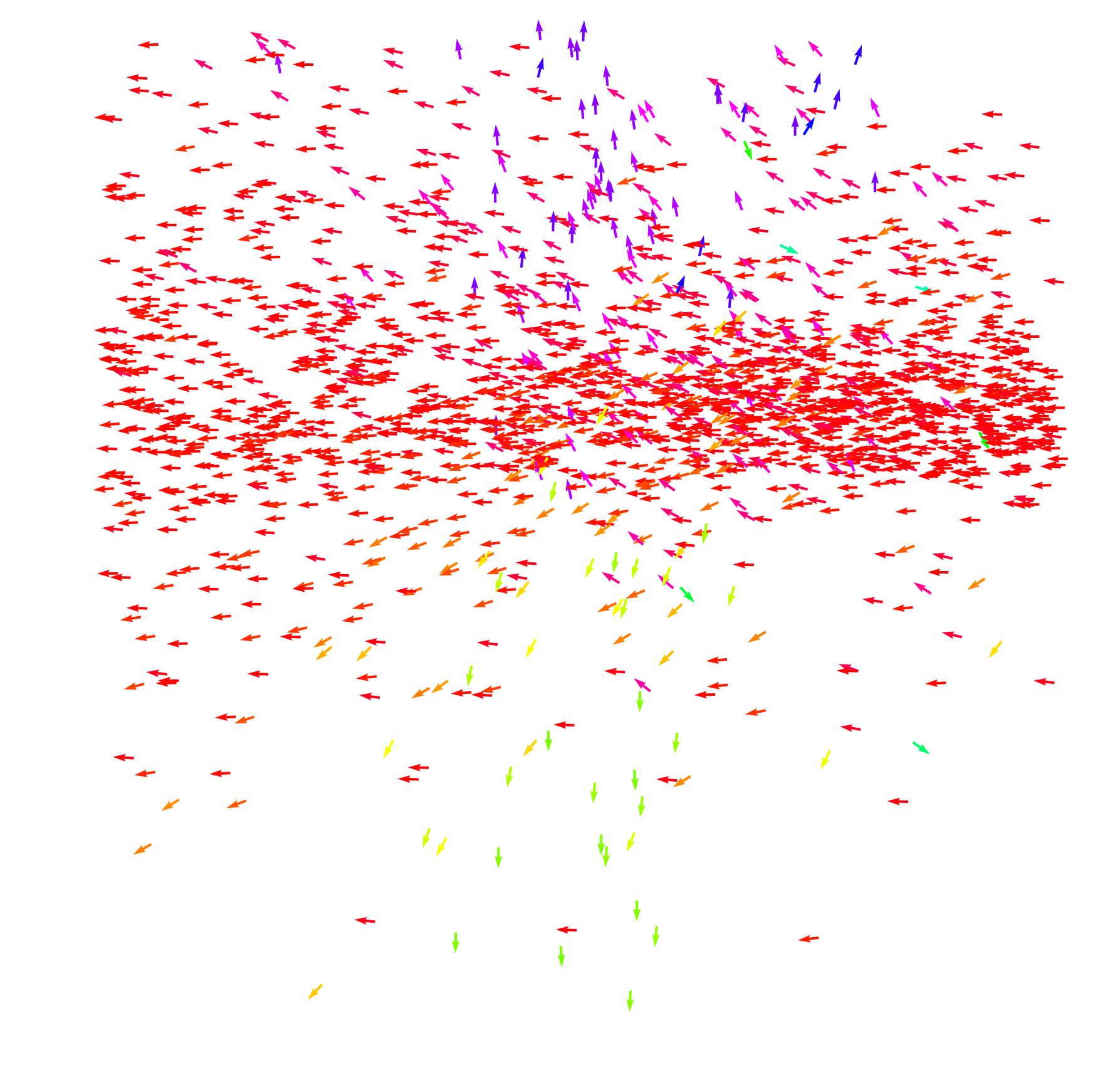}
    \label{fig:uic_highrise}}
  \caption{Illustration of the GPS data and the underlying noise for low and high-built areas of the UIC dataset. Both areas have the same dimension of $100\times100 $ square meters. The hue is based on the heading (best viewed in color).}
  \label{fig:uic_crop_noisy}
\end{figure}

Due to the noise from GPS sensors, locating road centerlines is difficult. Even in areas of low-built buildings, the noise from the sensors is at the same magnitude as the distance between adjacent roads. Thus, GPS locations recorded by the sensors might land outside of the road or even on the lane with the opposing traffic (see Figure \ref{fig:uic_lowrise}). 
This effect is exacerbated in areas of high-built buildings, with errors reaching 50\si{\metre} as shown in Figure \ref{fig:uic_highrise}. Worth noting is the imbalanced nature of the data with the west bound volume traffic being magnitude larger than the other directions. 
Thus, locating road centerlines is a hard task 
for which we have developed Iterated Trajectory Mean Shift (ITMS) in Section \ref{sec:improved}.

\subsection{Airport spatial complexity}\label{sec:airport_complex}


\begin{table}[b]
\caption{Details of UIC and FAA datasets.}
\label{tab:dataset_complexity}
\begin{tabular}{c|rr}
\toprule
\textbf{Attribute (unit)} & \multicolumn{1}{c}{\textbf{UIC}} & \multicolumn{1}{c}{\textbf{FAA}} \\ \hline \hline
\multicolumn{1}{c|}{Junction degrees} &
\multicolumn{1}{c}{3,4} & \multicolumn{1}{c}{3,4,5,6,7} \\ \hline
\multicolumn{1}{c|}{Number of Junctions} & \multicolumn{1}{c}{41} & \multicolumn{1}{c}{285} \\ \hline
\multicolumn{1}{c|}{Speed (\si[per-mode=symbol]{\metre\per\second})}
& {\begin{tabular}[r]{@{}r@{}}  8.9462 \\ $\pm$  3.5758 \end{tabular}}
& {\begin{tabular}[r]{@{}r@{}} 10.9225 \\ $\pm$ 13.3174 \end{tabular}} \\ \hline
\multicolumn{1}{c|}{\begin{tabular}[c]{@{}c@{}}Distance between points \\ in a trajectory (\si{\metre})\end{tabular}}
& {\begin{tabular}[r]{@{}r@{}} 24.4113 \\ $\pm$ 3.3066 \end{tabular}}
& {\begin{tabular}[r]{@{}r@{}} 19.3622 \\ $\pm$ 16.8242 \end{tabular}} \\ \hline
\multicolumn{1}{c|}{\begin{tabular}[c]{@{}c@{}}Distance between points \\ in a trajectory (\si{\second})\end{tabular}}
& {\begin{tabular}[r]{@{}r@{}} 3.6171 \\ $\pm$ 3.6758 \end{tabular}}
& {\begin{tabular}[r]{@{}r@{}} 5.8594 \\ $\pm$ 259.2738 \end{tabular}} \\ \bottomrule
\end{tabular}
\end{table}

As the FAA dataset only spans 24 hours of data recordings, only a few of trajectories have a complete spatial overlap of routes, that is, similar start and end points, as well as taking similar path along the taxiway while in transit. However, the spatial order of possible trajectory paths is still limited as airplanes have to follow the taxiway. Thus, although the mobility pattern within a day is highly irregular, the routine path travelled is not. This is only true for the airplane trajectories.

\begin{figure}[t]
  \subfigure[Histogram of trajectory lengths for the UIC dataset.]{\includegraphics[width=0.9\columnwidth]{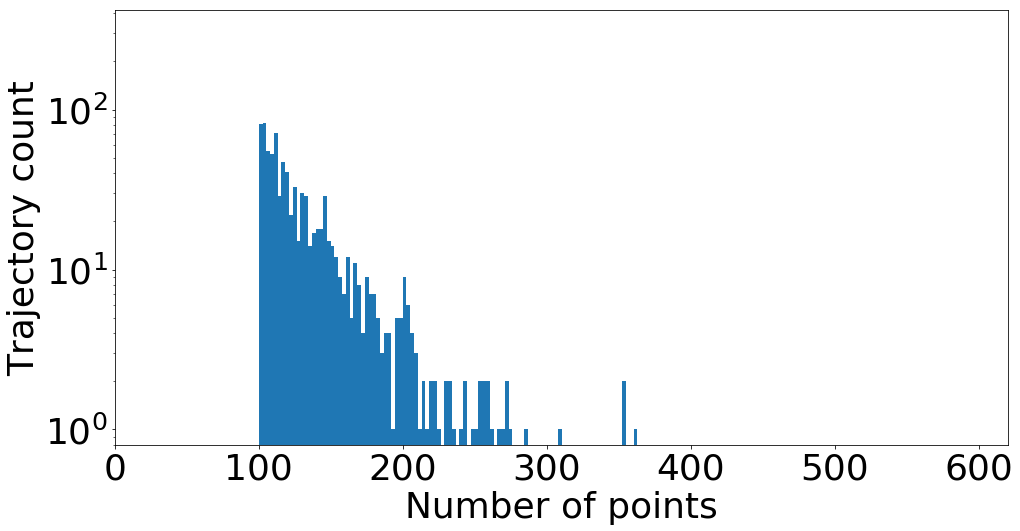}\label{fig:traj_len_UIC}}
  \subfigure[Histogram of trajectory lengths for the FAA dataset.]{\includegraphics[width=0.9\columnwidth]{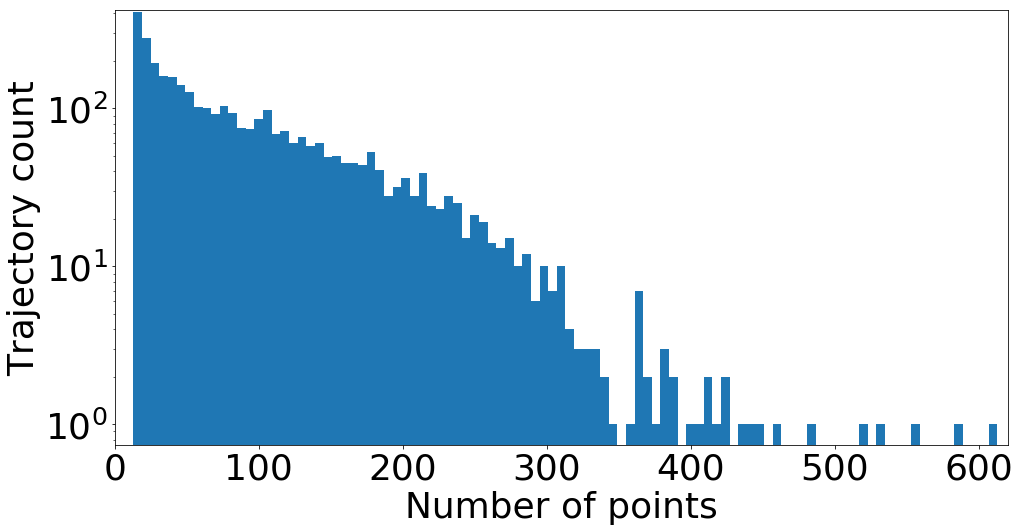}\label{fig:traj_len_FAA}}
  \caption{Histograms of trajectory lengths for both datasets.}
  \label{fig:traj_len}
\end{figure}

A significant portion of the trajectories are generated by the ground vehicles, which  deviate from  taxiways and take shortcuts, resulting in very long trajectories as shown in the distribution (Figure \ref{fig:traj_len}). This makes the dataset irregular, and is one of the main sources of noise which is unique to  airports.


Furthermore, this dataset has a high spatial complexity for a number of reasons. Firstly, it has a more complex geometry. The UIC dataset has junctions with degree 3 (T-junctions) and 4 (cross-junctions). In contrast, the junction degrees in the FAA dataset range between 3 and  6. For a quantitative overall comparison, the mean degree for junctions in the FAA dataset is 3.49, which higher than the value for UIC dataset, which is 3.37. Moreover, this dataset has nearly 7 times more junctions while occupying the similar area as UIC. Thus, junctions are closer to each other. For a quantitative comparison, the mean pairwise nearest junction distance for the airport is 33.7 m, which is much lower than 153 m for regular roads. The distance for the above analysis was computed as follows: for each junction, we find its nearest neighbor, we compute the distance, and we average over all possible such pairs.


All of these factors combined highlight the complexity of the dataset \eg, reflected by the higher standard deviations of nearly all of the attributes \ref{tab:dataset_complexity}, as well as a greater variation in the trajectory length (Fig. \ref{fig:traj_len}). In particular, the take-off and landing trajectory segments correspond to very high speeds, giving a positive skew to the distribution of speed and spatial distance between points within a trajectory.

\section{Definition and Problem Statement}
\label{sec:def}

\begin{table}[htbp]
\caption{Notations for  various attributes we access.}
\label{tab:notation}
\centering
\begin{tabular}{cc|c}
\textbf{Notation}             & \hspace{10pt} & \textbf{Description}  \\ \hline\hline
$\mathbf{p}_{(\mathbf{x})}; \mathbf{c}_{(\mathbf{x})}$ &  &  coordinates (2D) \\ \hline
$\mathbf{p}_{(h)}; \mathbf{c}_{(h)}$ &  &  heading (1D)                  \\ \hline
$\mathbf{p}_{(s)}; \mathbf{c}_{(s)}$ &  &  speed (1D)                     \\ \hline
$\mathbf{p}_{(t)}         $ &  & timestamp (1D)                \\ \hline
$\mathbf{p}_{(a)}         $ &  & altitude (1D)                \\ \hline
$\mathbf{p}_{(\kappa)}$&& is covered (boolean)\\ \hline
$\mathbf{c}_{(w)}         $ &  & weight (1D)                    \\ \hline
$\mathbf{c}_{(d)}         $ &  & degree (1D)                    \\ \hline
$\mathbf{c}_{(\overline{d})}         $ &  &  degree upper bound (1D)                    \\ \hline
$\mathbf{c}_{(l)}         $ &  & label (1D)                    \\ \hline
$|\mathcal{P}|; |\mathcal{T}|; |\mathcal{C}|; |\mathcal{M}|; |\mathcal{E}|$&  & set cardinality\\ \hline

$||\mathbf{a}-\mathbf{b}||_2$ where $\mathbf{a},\mathbf{b}$ are vectors 
&  & euclidean distance\\ \hline
{\begin{tabular}[c]{@{}c@{}}$\alpha(\mathbf{p}, \mathbf{q})$ 
\end{tabular}}
&  &
{\begin{tabular}[c]{@{}c@{}}heading difference\\(modulo operator)\end{tabular}}\\ 
\hline
\end{tabular}
\end{table}

In what follows, we define a GPS point $\mathbf{p} \in \mathbb{R}^5$ as a 5 dimensional vector consisting of a {\em latitude}, {\em longitude}, {\em speed}, {\em heading} and a {\em timestamp}. The timestamp is in the UNIX time format, which is the number of seconds elapsed since the midnight of 1 January 1970.

$\mathcal{P}^{(H)}$ is a set of $H$ points $\mathbf{p}$ so that $\mathcal{P}^{(H)}\equiv\{\mathbf{p_h}\}^{H}_{h=1}$. A trajectory $\mathbf{T}^{(I)}$ is an ordered sequence of $I$ points $\mathbf{p}$ so that $\mathbf{T}^{(I)} \equiv\{\mathbf{p_i}\}^{I}_{i=1}$. A set of trajectories $\mathcal{T}$ is a set of $J$ trajectory sequences $\mathbf{T}$ such that $\mathcal{T} \equiv \{\mathbf{T^{(I_j)}_j}\}^J_{j=1}$. As our notation suggests, a set of trajectories $\mathcal{T}$ may consist of trajectories of different lengths indicated by $I_j$.

In this paper, we are reducing a road into a sequence of line segments (null thickness), defined by centerline points $\mathbf{c}\in\mathcal{C}$ which  represent the road centerline. A road map $\mathcal{M}$ is represented as an annotated and directed graph 
with annotated centerline points $\mathcal{C}$ and 
edges $\mathbf{e}\in \mathcal{E}$ (from,to,weight) so that $\mathcal{M}\equiv \left(\mathcal{C},\mathcal{E}\right)$. 
%
%
In Table \ref{tab:notation}, we defined notations to access specific data attributes from the objects we have previously defined. The table also includes `intermediate' attributes required by our algorithm, such as {\em covered} and {\em weight}. All headings and angles are in radians. 

\begin{figure}[b]
  \subfigure[Input: trajectories from sensors]{\includegraphics[width=0.45\columnwidth]{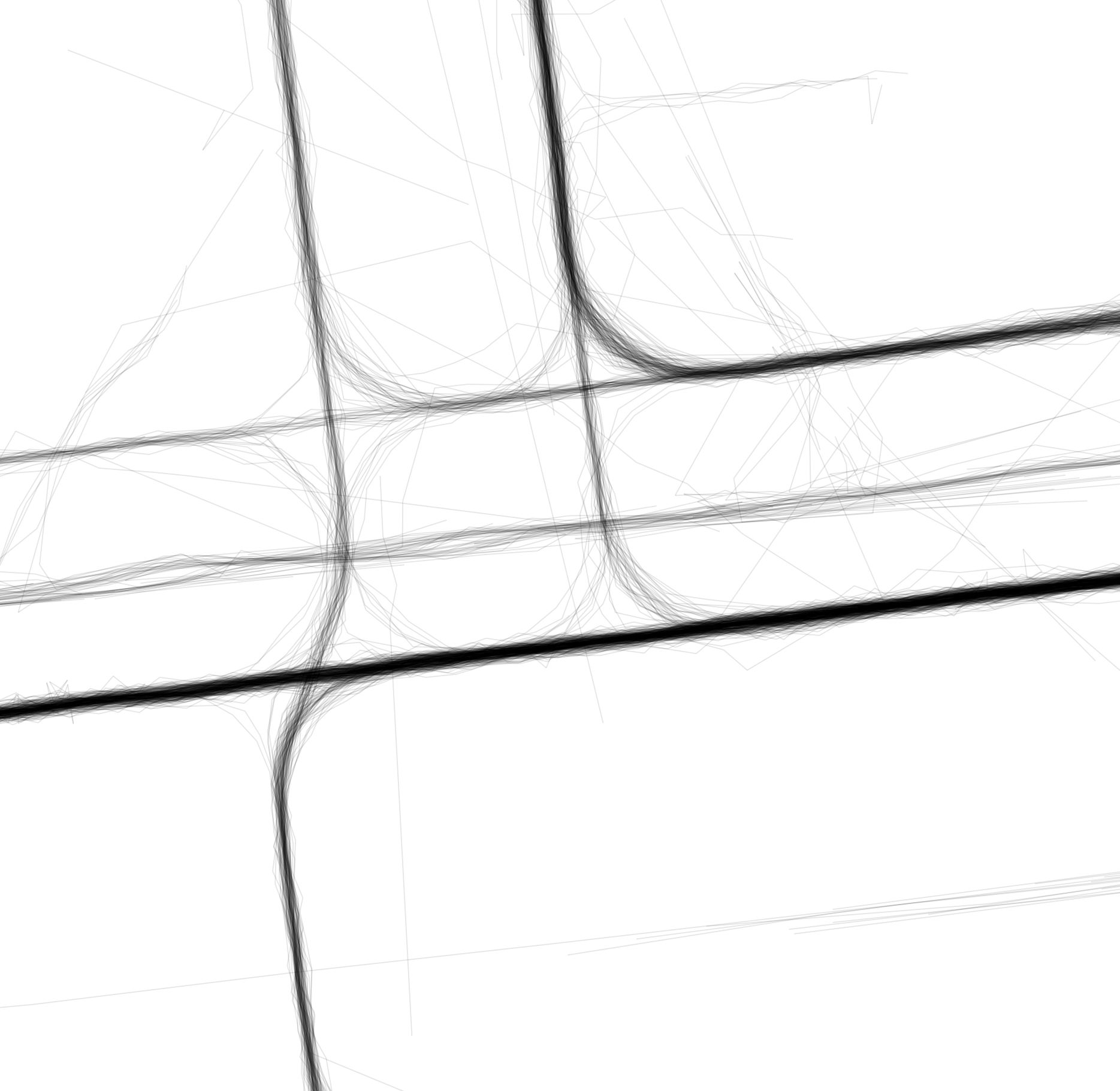}\label{fig:task_gps}}
\hspace{0.4cm}
  \subfigure[Output: Road Map]{\includegraphics[width=0.45\columnwidth]{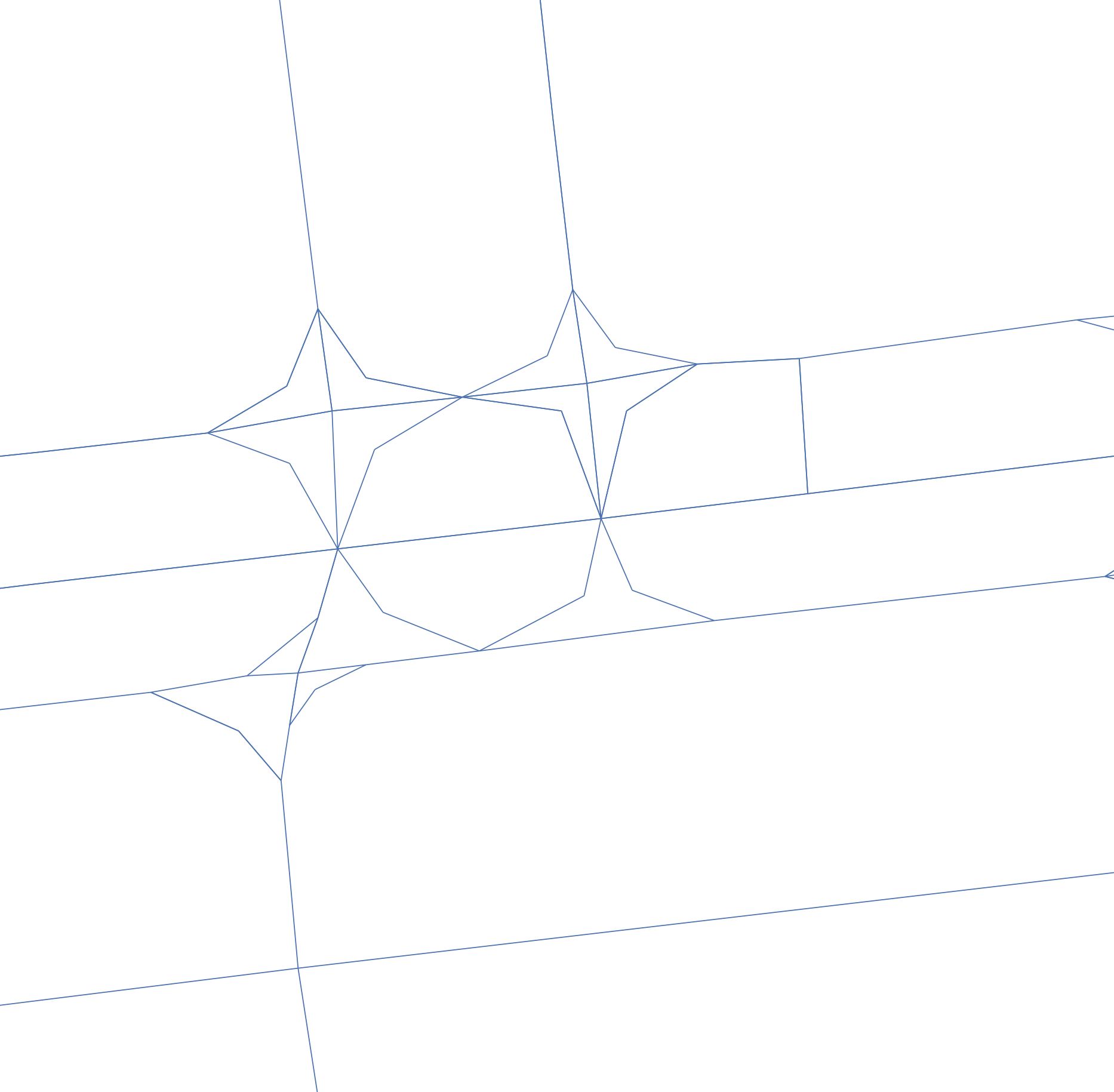}\label{fig:task_map}}
  \caption{Our COLTRANE takes  trajectories generated from sensors \ref{fig:task_gps} and produces the map \ref{fig:task_map}.}
  \label{fig:task}
\end{figure}

Our problem can be formalized as (i) inferring the road map $\mathcal{M}$ (\eg, Fig. \ref{fig:task_map}) from the set of trajectories $\mathcal{T}$ (\eg, Fig. \ref{fig:task_gps}), and (ii) detecting junctions to assign one of the 5 possible labels for every centerline point
$\mathbf{c}_{(l)}\forall\mathbf{c}\in\mathcal{C}$
, where
$\mathbf{c}_{(l)}$
can take on labels such as {\em not-a-junction}, {\em Y} or {\em T} junction, and {\em cross} or {\em star} intersection.

\section{Methodology}
\ap{Shall we change the section name to COLTRANE?}

\begin{figure*}[htb]
\centering
  \includegraphics[width=0.9\linewidth]{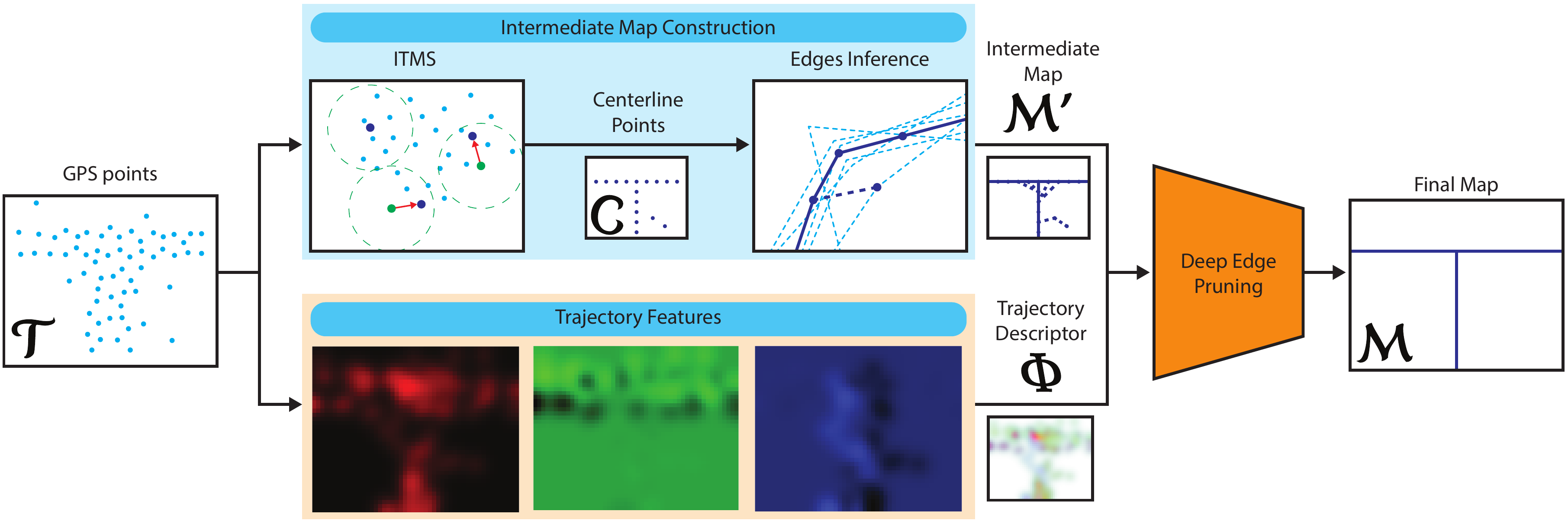}
  \caption{Our pipeline. Firstly, we generate an intermediate road map by clustering GPS trajectories using ITMS to approximate the centerline points and we connect them. In parallel, we construct a trajectory descriptor by merging trajectory features inferred from GPS datapoints. We then use the trajectory descriptor as an input to CNN to determine the degree 
  and junction type for each centerline point. After pruning, we obtain the road map.}
\label{fig:overall_arch}
\end{figure*}

In what follows, we explain our COLTRANE approach. We proceed by constructing an intermediate road map $\mathcal{M'}$ from the set of trajectories $\mathcal{T}$ generated by the sensors.

As $\mathcal{M'}$ contains many false positive edges, it is not trivial to remove them. The difficulty of this step is shown in Figure \ref{fig:uic_lowrise}. All of the east-to-west traffic joining the intersection is directed south bound, while all the east-to-west traffic past the intersection is coming from the the north-to-south traffic. A less sophisticated algorithm would mistakenly form an east-to-west edge.

In order to trim false positives, we develop a novel  trajectory descriptor $\mathbf{\Phi}$, which we use as an input to a  Convolutional Neural Network (CNN), a highly discriminative model which aids  complex edge pruning. The output of the edge pruning module forms the final route map $\mathcal{M}$. Figure \ref{fig:overall_arch} gives an overview of our pipeline.

\subsection{Intermediate Map Construction}\label{sec:intermediate_map}

The process of generating the intermediate map $\mathcal{M'}$ can be divided into two steps: (i) extracting centerline points $\mathbf{c}\in\mathcal{C}$ and (ii) inferring the edges $\mathbf{e}\in\mathcal{E}$ that connect those points. The centerline points $\mathbf{c}\in\mathcal{C}$ are extracted using ITMS. Then, our algorithm uses trajectories $\mathcal{T}$ to infer directional links between the centerline points $\mathbf{c}\in\mathcal{C}$ to form a set of edges $\mathcal{E}$. As the false positive edges will be pruned in the next step, our algorithm forms all the plausible edges first. Centerline points and directed edges form a directed graph representing the intermediate map $\mathcal{M'}=\left(\mathcal{C},\mathcal{E}\right)$.

\subsubsection{Iterated Trajectory Mean Shift Sampling (ITMS)}
\label{sec:improved}

Traj-Mean Shift,  proposed in \cite{chen2016city},  locates road centerlines. It builds on the mean shift clustering algorithm by shifting cluster centers towards the mean location of their neighbouring points. The two main properties of Traj-Mean Shift are: (i) cluster centers shift along an axis perpendicular to the heading, and (ii) the histogram takes heading and speed into account. Since Traj-Mean Shift locates road centerlines, we will refer to the cluster centers as centerline points. 

\begin{algorithm}[t]
\caption{Iterated Trajectory Mean Shift Sampling (ITMS).}
\vspace{-0.3cm}
\label{algo}
\SetAlgoLined
\SetKwInOut{Input}{input}\SetKwInOut{Output}{output}
\Input{$\mathcal{P}$ - a set of GPS points}
\Output{$\mathcal{C}$ - a set of centerline points}
set $converged=False$\;
\While{$\neg$ converged}{
    set $\mathbf{p}_{(\kappa)}=0:
        \forall\mathbf{p}\in\mathcal{P}$;
        set $\mathcal{C'}=\emptyset$
        \tcp*{initialization}\
    \While(\tcp*[f]{while  $\mathbf{p}\in\mathcal{P}$ are not covered})
    {$\neg\mathbf{p}_{(\kappa)}=1, \forall\mathbf{p}\in\mathcal{P}$}{
        \tcp{pick a random point and initialize with it the road centerline point}
       pick randomly $\mathbf{p}\in\mathcal{P}:
       \mathbf{p}_{(\kappa)}=0$, set $\mathbf{c}_{(\mathbf{x},s,h)} =\mathbf{p}_{(\mathbf{x},s,h)}$,
       set $m=\infty$\;
        \While(\tcp*[f]{while movement of $\mathbf{c}>\epsilon$})
        {$m > \epsilon$}{{
            \tcp{set similar neighboring points as covered}
            set $\mathbf{n}_{(\kappa)}=1,
                \forall\mathbf{n}\in\mathcal{N}\equiv
                \mathbf{n}\in\mathcal{P}: ||\mathbf{c}-\mathbf{n}||_2 \leq\gamma $\\
 \Indp\Indp\Indp\Indp\Indp\Indp\Indp\Indp$\;\land\;\alpha(\mathbf{c},\mathbf{n})<\eta $\;}
            \tcp{find the mean}
            construct $\mathbf{histogram}$ at $\mathbf{c}$ and perpendicular to $\mathbf{c}_{(h)}$\ with $\beta$ bins\;
            project $\mathbf{n}$ to $\mathbf{histogram}$ weighted according to $w(\mathbf{c},\mathbf{n}; \mathcal{N}), \forall\mathbf{n}\in\mathcal{N}$\;
            set $\mathbf{mean\_bin}=$ the bin containing the mean value of $\mathbf{histogram}$*\;
            set $\mathcal{U}\equiv\{
                \mathbf{u}\in\mathcal{N}:
                \mathbf{u}\in\mathbf{mean\_bin}
                \}$\;
            \tcp{shift the road centerline point $\mathbf{n}$ to the mean}
            set $m=||\mathbf{mean\_bin}-\mathbf{c}||_2$; \tcp{how far $\mathbf{c}$ has moved}

            set $\mathbf{c}_{(\mathbf{x})}=\mathbf{mean\_bin}_{(\mathbf{x})}$; 
            
            set $\mathbf{c}_{(s)}=
                \frac{1}{|\mathcal{U}|}\sum_{\mathbf{u}\in\mathcal{U}}\mathbf{u}_{(s)}\cdot\upsilon(\mathbf{c},\mathbf{u})$; 
                
                set $\mathbf{c}_{(h)}=
             \frac{1}{|\mathcal{U}|}\sum_{\mathbf{u}\in\mathcal{U}}\mathbf{u}_{(h)}\cdot\upsilon(\mathbf{c},\mathbf{u})$;

            \tcp{set similar neighboring points as covered after the shifts}
            set $\mathbf{n}_{(\kappa)}=1, 
 \forall\mathbf{n}\in\mathcal{N}\equiv\mathbf{n}\in\mathcal{P}:||\mathbf{c}-\mathbf{n}||_2\leq\gamma$\\
 \Indp\Indp\Indp\Indp\Indp\Indp\Indp\Indp$\;\land\;\alpha(\mathbf{c},\mathbf{n})<\eta$\;
        }
        set $\mathcal{C'}=\mathcal{C'}\cup\{\mathbf{c}\}$;  \tcp{append $\mathbf{c}$ to $\mathcal{C'\!}$}
    }
    \leIf{$|\mathcal{P}|=|\mathcal{C'}|$}{
        set $converged = True$
    }{
        set $\mathcal{P}=\mathcal{C'\!}$
    }
}
set $\mathcal{C}=\mathcal{C'\!}$;
\end{algorithm}

The algorithm  proceeds as follows. Firstly, all points are set as {\em uncovered} \ie, $\mathbf{p}_{(\kappa)}\!=\!0, \forall\mathbf{p}\!\in\!\mathcal{P}$. Then,  a random point is chosen as an uncovered point $\mathbf{p}\!\in\!\mathcal{P}:\mathbf{p}_{(\kappa)}\!=\!0$ that is used to initialize the centerline point by setting $\mathbf{c}_{(\mathbf{x},s,h)} \!=\!\mathbf{p}_{(\mathbf{x},s,h)}$.
Then, neighbors $\mathcal{N}$ of the centerline point $\mathbf{c}\in\mathcal{C}$, defined as $\mathcal{N}\equiv\{\mathbf{n}\in\mathcal{P}: ||\mathbf{c}-\mathbf{n}||_2 \leq \gamma\}$ are formed ($\gamma$ is a constant). Neighbours $\mathbf{n}\in\mathcal{N}$ are projected onto a weighted histogram that is centered at $\mathbf{c}_{(x)}$ and perpendicular to $\mathbf{c}_{(h)}$. The weights for the GPS point $\mathbf{p}\in\mathcal{P}$ in the histogram for the centerline point $\mathbf{c}\in\mathcal{C}$ are set as follows:
\begin{equation}
    w(\mathbf{c},\mathbf{n}; \mathcal{N})=
    \frac{
    \exp\big(-\!{\alpha^2\big(\mathbf{c},\mathbf{n}\big)}/{2\sigma^2_h}\big)\exp\big(-\!{\big(\mathbf{c}_{(s)}\!-\!\mathbf{n}_{(s)}\big)^2}/{2\sigma^2_s}\big)
    }{\sum\limits_{\mathbf{n'}\!\in\mathcal{N}}\!
    \exp\big(-\!{\alpha^2\big(\mathbf{c},\mathbf{n'}\big)}/{2\sigma^2_h}\big)\exp\big(-\!{\big(\mathbf{c}_{(s)}\!-\!\mathbf{n'}_{(s)}\big)^2}/{2\sigma^2_s}\big)}, 
    \label{eq:bin_weight}
\end{equation}
where $\sigma_h$ and $\sigma_s$ are RBF radii. The centerline point $\mathbf{c}\in\mathcal{C}$ is then shifted to the coordinate of the histogram bin with the largest value, denoted as $\mathbf{c'}$ from here on, and assigned a weight equal to the number of its new neighbours 
$\mathbf{c'}_{(w)} =|\mathcal{N'}|:\mathcal{N'}\equiv\{\mathbf{n'}\in\mathcal{P}: ||\mathbf{c'}-\mathbf{n'}||_2 \leq\gamma\}$.
Finally, all new neighbours are set as covered $\mathbf{n'}_{(\kappa)}\!=\!1, \forall\mathbf{n'}\in\mathcal{N'}$.


\begin{figure}[b]
\vspace{-0.5cm}
    \centering
    \subfigure[Traj-Mean Shift \cite{chen2016city}]{
        \includegraphics[width=0.45\linewidth]{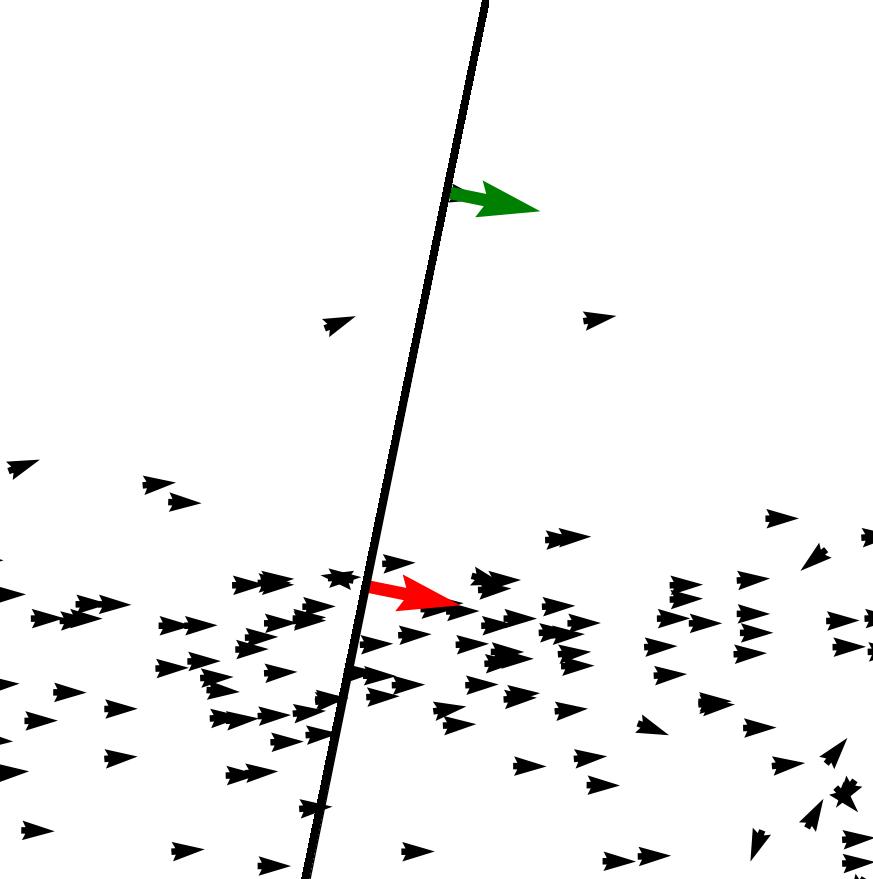}\label{fig:chen_b}}
    \hfill
    \subfigure[ITMS]{
        \includegraphics[width=0.45\linewidth]{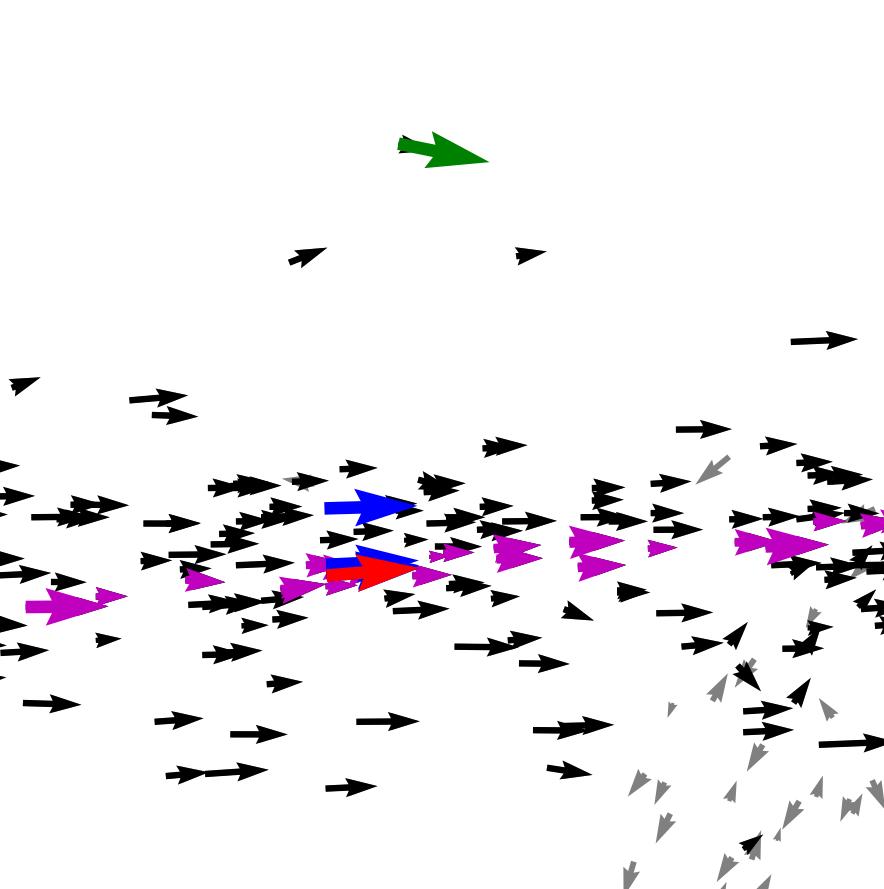}\label{fig:dmi_b}}
    \caption{An illustrated comparison of Traj-meanshift and ITMS (best viewed in color). Green arrows are the initial points and red arrows are the final centerline points. The thick black line is the axis of the histogram (see Section \ref{sec:improved}). Black arrows are the neighbours, while gray arrows are not. Blue arrows are the intermediate centerline points. Magenta arrows are the neighbours that fall into the mean bin. See Algorithm \ref{algo} for more details. 
}
\end{figure}

Figure \ref{fig:chen_b} illustrates that Traj-Mean Shift \cite{chen2016city} fails to correctly determine the location and heading of the road centerline if the initial point $\mathbf{p}\in\mathcal{P}$ is too far from the actual road centerline. Note that the final centerline point has the wrong heading as well. This is problematic for more complex/noisy GPS data \eg, from the airport tarmac as shown in Section \ref{sec:noisy_gps}.

In Algorithm \ref{algo}, to address this shortcoming, we propose Iterated Trajectory Mean Shift (ITMS), an extension of Traj-Meanshift with  improvements described below.

Firstly, the sample point $\mathbf{c}\in\mathcal{C}$ is shifted to the $\mathbf{mean\_bin}$ (if non-empty)
of the weighted histogram rather than the bin with the maximum value. Otherwise, the bin with maximum value is chosen  (note `*' in Algorithm \ref{algo}).
Secondly, to approximate well the speed and heading of $\mathbf{c}$, we adjust them by taking into account the speed and heading of other points in the selected bin: $\mathbf{c}_{(s)}=
                \frac{1}{|\mathcal{U}|}\sum_{\mathbf{u}\in\mathcal{U}}\mathbf{u}_{(s)}\cdot\upsilon(\mathbf{c},\mathbf{u}),\;
                \mathbf{c}_{(h)}=
             \frac{1}{|\mathcal{U}|}\sum_{\mathbf{u}\in\mathcal{U}}\mathbf{u}_{(h)}\cdot\upsilon(\mathbf{c},\mathbf{u}) ,$ where $\mathcal{U}\equiv\{
                \mathbf{u}\in\mathcal{N}:
                \mathbf{u}\in\mathbf{mean\_bin}
                \}$. Our weighting is defined as follows:
\begin{equation} \label{eq:weight_mean_heading}
    \upsilon(\mathbf{c},\mathbf{u})=1-\frac{\alpha(\mathbf{c}_{(h)},\mathbf{u}_{(h)})}{\pi}.
\end{equation}
Thirdly, only neighbors of $\mathbf{p}$, $\mathbf{n}$, and all intermediate $\mathbf{n}$ whose heading is close to their respective points, are considered as processed (black in Figure \ref{fig:dmi_b}) while neighbors with different headings are not considered processed (gray in Figure \ref{fig:dmi_b}). Finally,  Algorithm \ref{algo}  iterates the shifts until  $m\!\leq\!\epsilon$ and the set cardinality  $|\mathcal{C}'|$ does not change. 
%
%
We set the hyperparameters as follows: $\sigma_h=1.7$, $\sigma_s=13$, $\epsilon=1.5\si{\metre}$, $\gamma=15\si{\metre}$, $\eta=0.3\pi$, and $\beta=50$ .

\subsubsection{Edge Inference}

This module infers the intermediate set of directed edges $\mathcal{E}'$ between the centerline points from set $\mathcal{C}$ based on the trajectories $\mathcal{T}$. Because we use a powerful learner to prune the false positive edges in the next step, we form all plausible edges in this step. 
Since the the centerline points $\mathbf{c}\in\mathcal{C}$ were obtained through clustering, we  assume that every GPS point in $\mathcal{P}$ corresponds to some cluster  in $\mathcal{C}$. Thus, we  map every single GPS point in $\mathcal{P}$ to a centerline point in $\mathcal{C}$  by finding the centerline point $\mathbf{c}$ that minimizes expression  $s(\mathbf{c},\mathbf{p})\!=\!\|\mathbf{c}-\mathbf{p}\|_2 \cdot \alpha(\mathbf{c}, \mathbf{p})$. 
%
%
%
%
Specifically, for every pair of adjacent GPS points within a trajectory
$(\mathbf{p_i},\mathbf{p_{i+1}}):
\mathbf{p_i},\mathbf{p_{i+1}}\!\in\!\mathbf{T}\!\in\!\mathcal{T}$, we form a directed edge with the weight of 1,
$\mathbf{e}\!=\!(\mathbf{c_j},\mathbf{c_k},1)$,
where
$\mathbf{c_j}\!=\!\argmin{\mathbf{c}\in\mathcal{C}}s(\mathbf{c},\mathbf{p}_i)$ and 
$\mathbf{c_k}\!=\!\argmin{\mathbf{c}\in\mathcal{C}}s(\mathbf{c},\mathbf{p}_{i+1})$.
If $\mathbf{c_j}\!=\!\mathbf{c_k}$,  no edge is made. If an edge $\mathbf{e}\!=\!(\mathbf{c_j},\mathbf{c_k},w)$ with weight $w$ already exist,
%
 we increment the weight of that edge by 1, that is 
$\mathbf{e}\!=\!(\mathbf{c_j},\mathbf{c_k},w+1)$.
%
%
The set of all  intermediate edges  $\mathcal{E}'\!$ combined with the centerline points $\mathcal{C}$ from the previous section form the intermediate road map $\mathcal{M}'.$

\subsection{Trajectory Features and Traj. Descriptor}\label{ref:descriptor}

Below, we introduce our novel trajectory features extracted from $\mathcal{P}$ that capture spatial and velocity information. When combined, these trajectory features form the trajectory descriptor $\mathbf{\Phi}$, which is  an input to the CNN (described in the next section) which is able to learn spatial relationships between datapoints. Although trajectories are sequences of GPS points, spatial information is more important for the map inference than the  the sequential ordering of the GPS points. As CNN is a natural choice for feature map inputs, we design our trajectory descriptor to be a feature map. 



The trajectory features are binned into weighted 2D histograms to form a feature map (array). The dimension of each bin corresponds to $1\si{\metre}\!\times\!1\si{\metre}$ area. We then combine the histograms into a descriptor $\mathbf{\Phi}$ (an image/feature map with three channels).

The first trajectory feature is simply a binarized 2D histogram \cite{davies2006scalable}. For each bin in the histogram, it  set the value equal 1 if there is at least one GPS point that falls in it. Otherwise, the bin is set to 0.
Then, we extrapolate the speed and heading of each GPS point $\mathbf{p} \in \mathcal{P}$ and compute the $x-$ and $y-$directional velocities for each $\mathbf{p}\in \mathcal{P}$ respectively. 
For the second and third trajectory means, each histogram bin is computed by aggregating over $x-$ and $y-$directional velocities of $\mathbf{p}$ that fall into that bin, respectively. These last two channels are normalized within the $[0;1]$ range. 

\subsection{Deep Edge Pruning}

The intermediate map from Section \ref{sec:intermediate_map} contains false positive edges. 
Thus, we propose  a powerful CNN framework to prune edges.

Firstly, the trajectory descriptor $\mathbf{\Phi}$ described in Section \ref{ref:descriptor} is the input to our CNN which simultaneously performs the junction detection and classification to determines the degree for each centerline point $\mathbf{c}_{\bar{d}}\!:\mathbf{c}\in\mathcal{C}$. Finally, for each centerline point $\mathbf{c}\in\mathcal{C}$, we prune the edges $\mathbf{e}\in\mathcal{E}'$ in the intermediate map $\mathcal{M}'\!$ based on the weight of  edges. The pruned map is the final inferred road map $\mathcal{M}$.


\begin{figure}[t]
\centering
  \includegraphics[width=1\columnwidth]{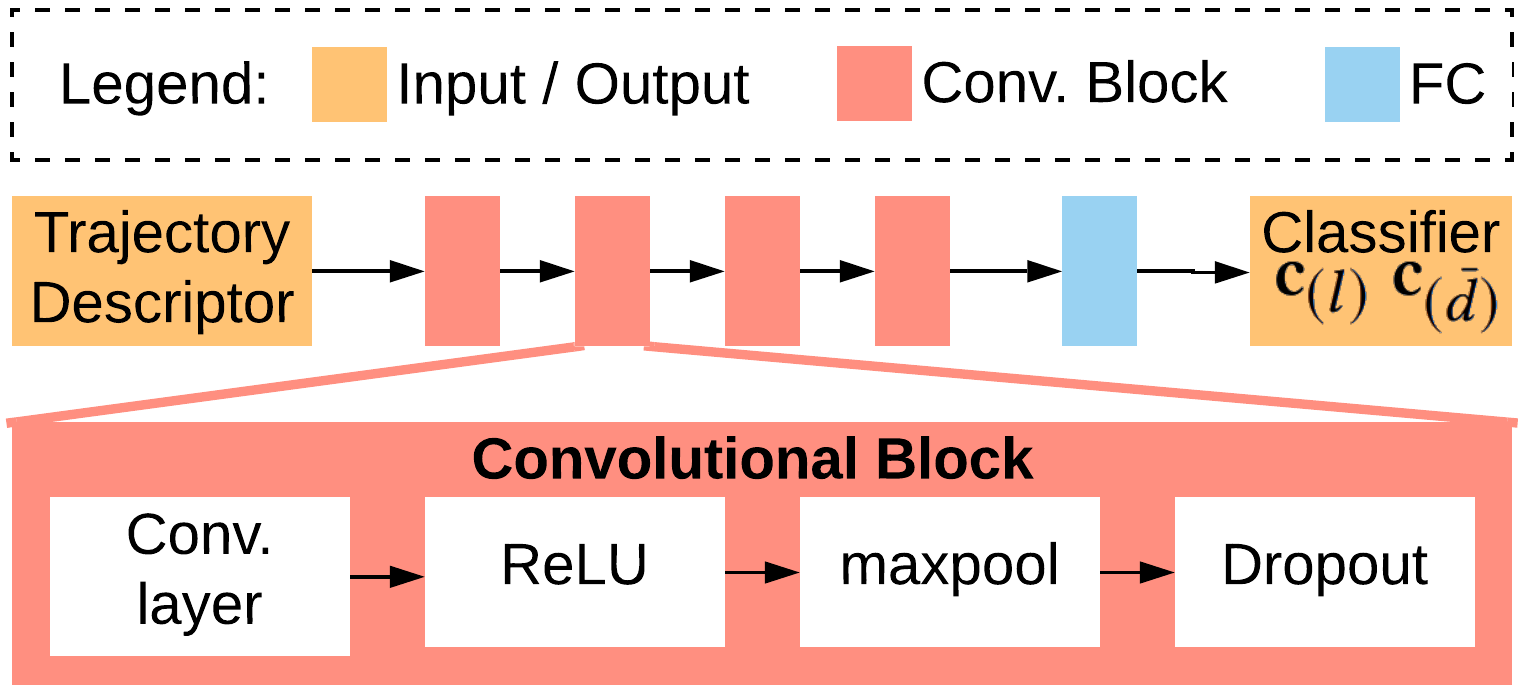}
  \caption{The architecture of our CNN  consist of four convolutional units  and a fully connected unit followed by the output layer. The convolutional unit consist of a convolutional layer, a Rectified Linear Unit (ReLU) activation function \cite{nair2010rectified}, maxpooling and dropout.}
  \label{fig:arch_cnn}
\end{figure}

\subsubsection{Junction Detection and Classification using CNN}

For each $\mathbf{c}\!\in\!\mathbf{C}$, we extract a patch from the trajectory descriptor $\mathbf{\Phi}$, centered at $\mathbf{c}_{(x)}$,  and we feed it to CNN. The patch size is $100\!\times\!100$ bins ($\times\!3$). %
The CNN resembles the VGG pipeline \cite{simonyan2014very} due to its simplicity and success in computer vision tasks. However, we used fewer layers as the most salient features should already be captured by our trajectory descriptor $\mathbf{\Phi}$. 
Thus, our CNNs consists of four convolutional units and a fully connected layer followed  by the output layer. Convolutional units consist of a convolutional layer with 32 filters of $3 \times 3$ size, the stride is $1 \times 1$. We use a Rectified Linear Unit (ReLU) as the activation function \cite{nair2010rectified}. The convolutional layer is followed by a $2 \times 2$ maxpooling and dropout. The fully connected layer has 512 filters. Figure \ref{fig:arch_cnn} shows the architecture of our network. We use the Adam optimizer with parameters taken from the original paper \cite{kingma2014adam}. We train the model for 100 epoch, with a batch size of 156. 
The output describes each centerline point by its degree (upper bound) $\mathbf{c}_{(\bar{d})}$ and a label $\mathbf{c}_{(l)}$ to detect and classify junctions, as described in Section \ref{sec:def}. 

\subsubsection{Edge Pruning}

Based on the degree inferred by the CNN, we prune the false positives from the intermediate edges $\mathcal{E}'$, leaving  the true positives. For every centerline point $\mathbf{c}\in\mathcal{C}$, we prune the edges, starting from the edge with a lowest edge weight, until all centerline points have degrees that are smaller or equal to the upper bound degree, that is $\mathbf{c}_{(d)}\leq\mathbf{c}_{(\bar{d})}$. The resulting set of edges $\mathcal{E}$ and the road centerlines $\mathcal{C}$ form the final road map $\mathcal{M}=(\mathcal{C},\mathcal{E})$.

\section{Results and Analysis}

Below, we compare our COLTRANE to Chen \etal \cite{chen2016city}. 
We describe the datasets we use, our experimental setup and the results.

\subsection{Experimental Setup}
For the junction detection and classification, we applied oversampling to account for the unbalanced classes (\eg, most instances belong to {\em road segment} classes), $8\times$ data augmentation (left-right flip and 4 rotations). The models were trained using 70\% of the data while the remaining 30\% was kept for testing \cite{chang2011libsvm,chen2016city}. The accuracy, Macro--F1 score, and confusion matrices are evaluated on the test set. We do not fine-tune any hyperparameters of our algorithm. For the purpose of visual evaluation in Figure \ref{fig:viz_eval}, the model was trained on the training set and predictions were made on the entire dataset.

Since we perform the multiclass classification with C classes, we use Macro--F1 score defined as:
$$\textrm{Macro--F1}=\sum_{c=1}^{C}\frac{tp_c}{2tp_c+fp_c+fn_c},$$
where $tp_c$, $fp_c$, $fn_c$ denote the true positive, false positive, and false negative, for class $c$, respectively.

\subsection{Empirical Evaluation Metric}
For evaluating the quality of our map inference algorithm, we used a graph matching method called `marble and holes' \cite{biagioni2012inferring}. From a starting location, the evaluation algorithm will traverse the inferred/ground truth map, dropping a marble/hole at regular interval $d$, until it is $r$ meters away from the starting point. If the distance between a marble and a hole is less than $d$, they are considered as a match. Unmatched marbles and holes are considered false positives and negatives, respectively. Thus, precision, recall and F1 scores can be calculated. For the UIC dataset, we fixed the sampling rate to $r\!=\!100$m, and varied the matching distance $d$ between 1m and 30m \cite{chen2016city}.

\subsection{Map Inference}

\begin{figure}[b]
\centering
  \subfigure[UIC]{
    \includegraphics[width=0.47\columnwidth]{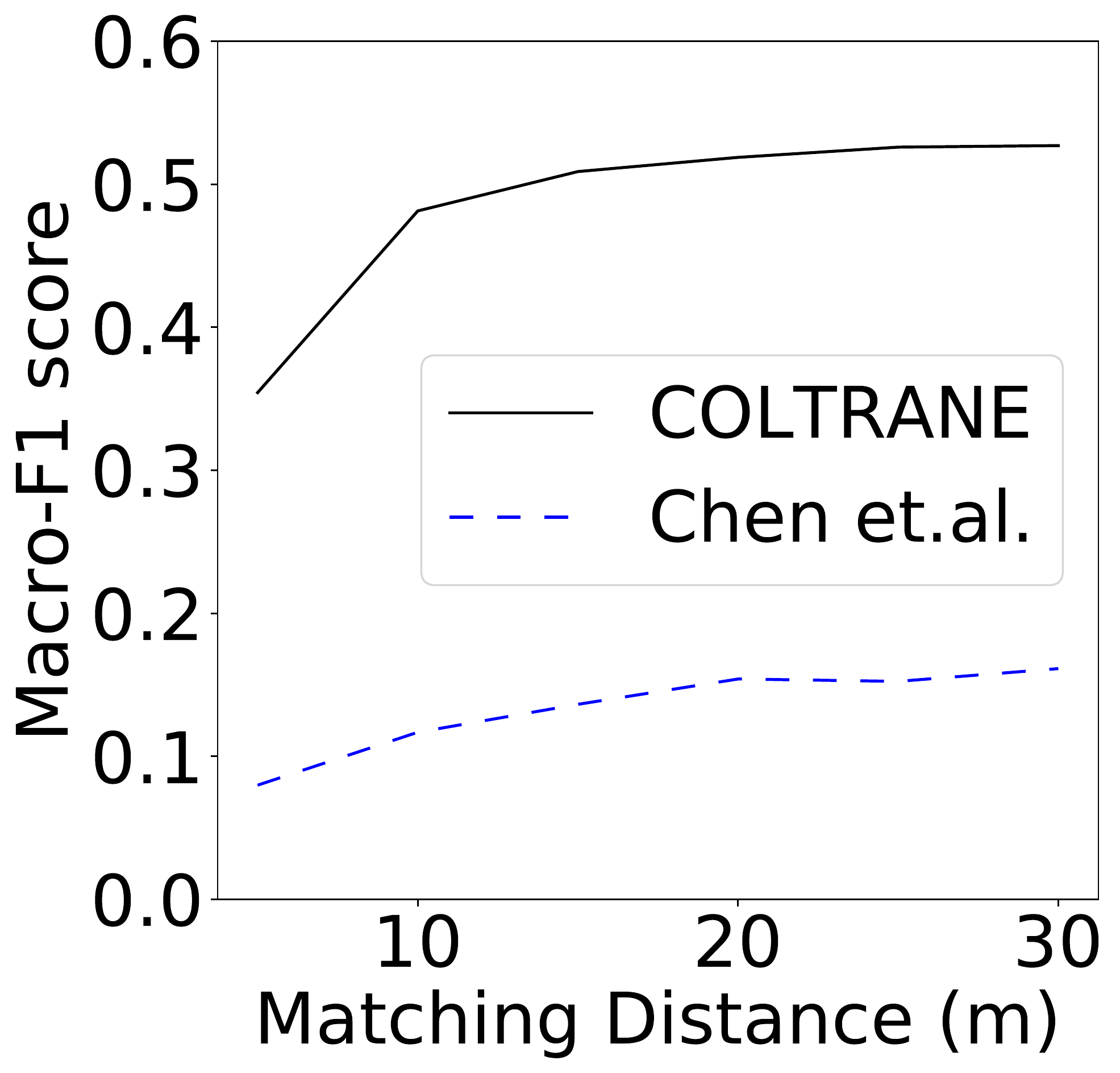}}
  \hspace{0.01\columnwidth}
  \subfigure[FAA]{
    \includegraphics[width=0.47\columnwidth]{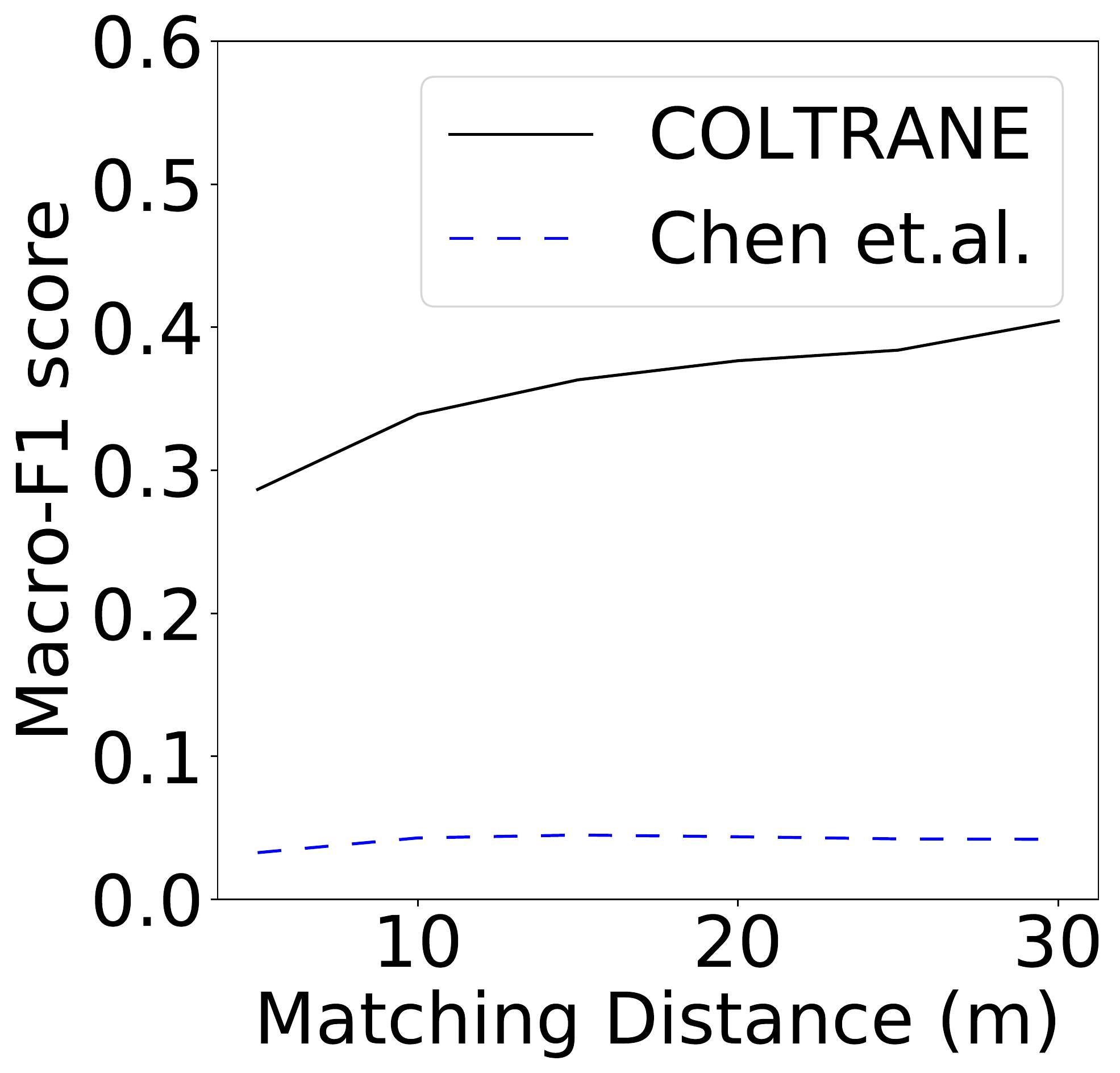}}
  \caption{Performance comparison (Macro--F1 score) of the map inference between COLTRANE and the approach of Chen \etal \cite{chen2016city} for a varying matching distance \cite{biagioni2012inferring}.}
  \label{fig:mh}
\end{figure}

Figure \ref{fig:mh} shows that COLTRANE attains the best Macro--F1 results across all matching distances for both datasets. The improvement ranges from 27\% on the UIC dataset with the matching distance of 5 \si{\metre}, to 37\%, also on the UIC dataset given the matching distance of 25 \si{\metre}. Moreover, these results also confirm our findings in Section \ref{sec:airport_complex} that the FAA dataset (airport data) is spatially more complex, thus posing a challenge for the map inference.

\subsubsection{Visual Evaluations}

\begin{figure}[t]
    \centering
    \subfigure[COLTRANE on UIC]{
        \includegraphics[width=0.48\columnwidth]{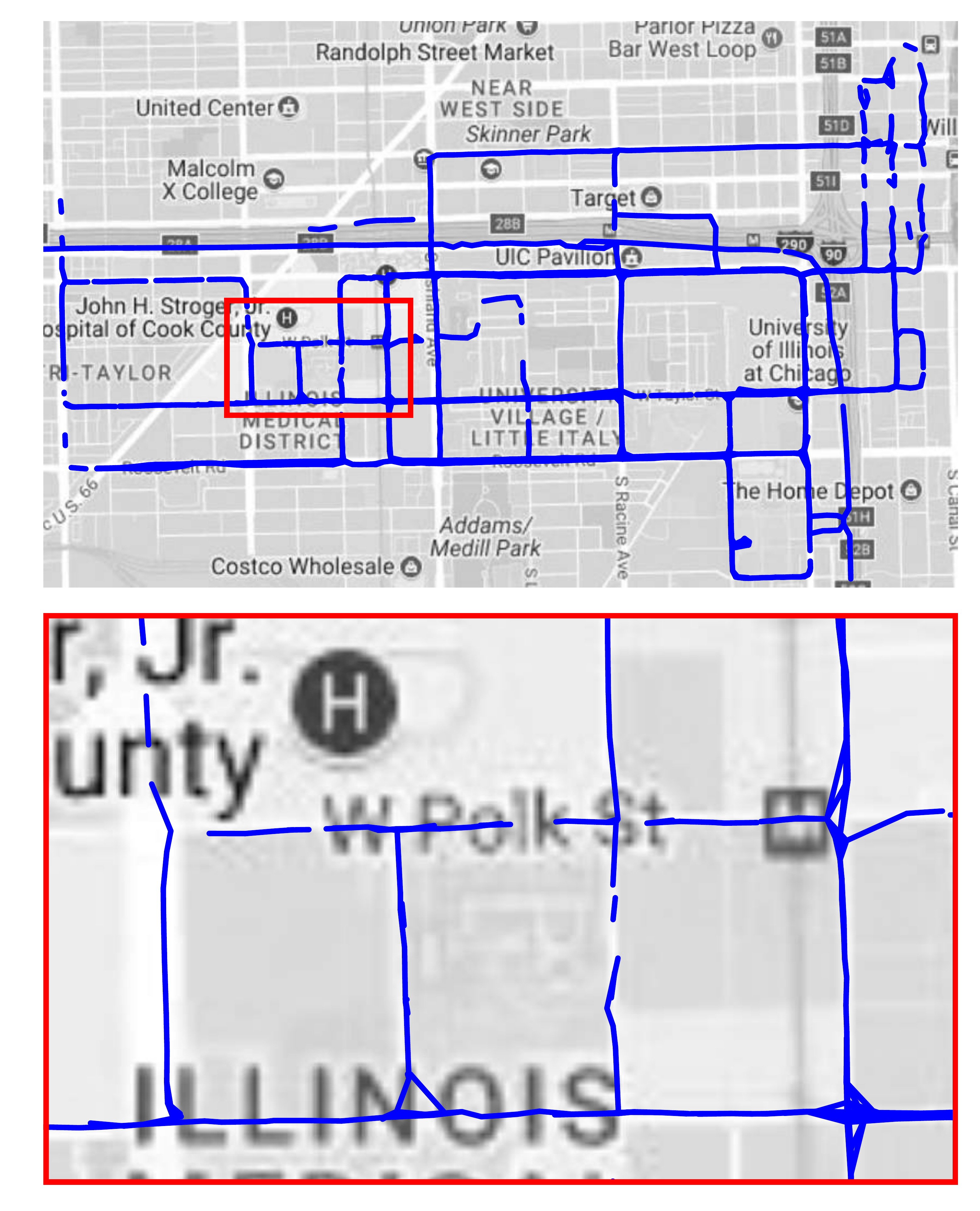}\label{fig:uic_results_a}}
    \subfigure[Chen \etal \cite{chen2016city} on UIC]{
        \includegraphics[width=0.48\columnwidth]{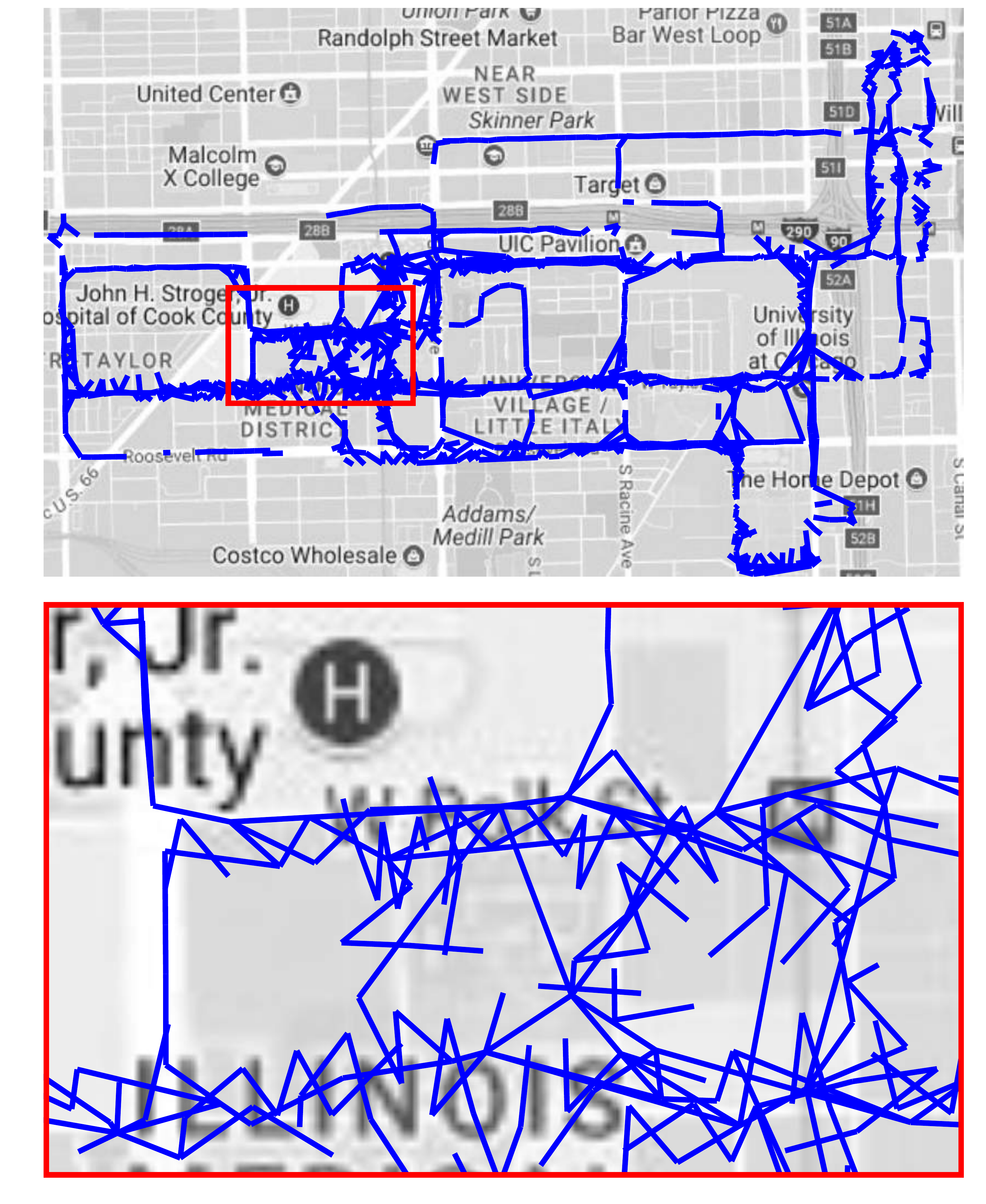}\label{fig:uic_results_c}}
    \\ \vspace*{-3ex}
        \subfigure[COLTRANE on FAA]{
        \includegraphics[width=0.48\columnwidth]{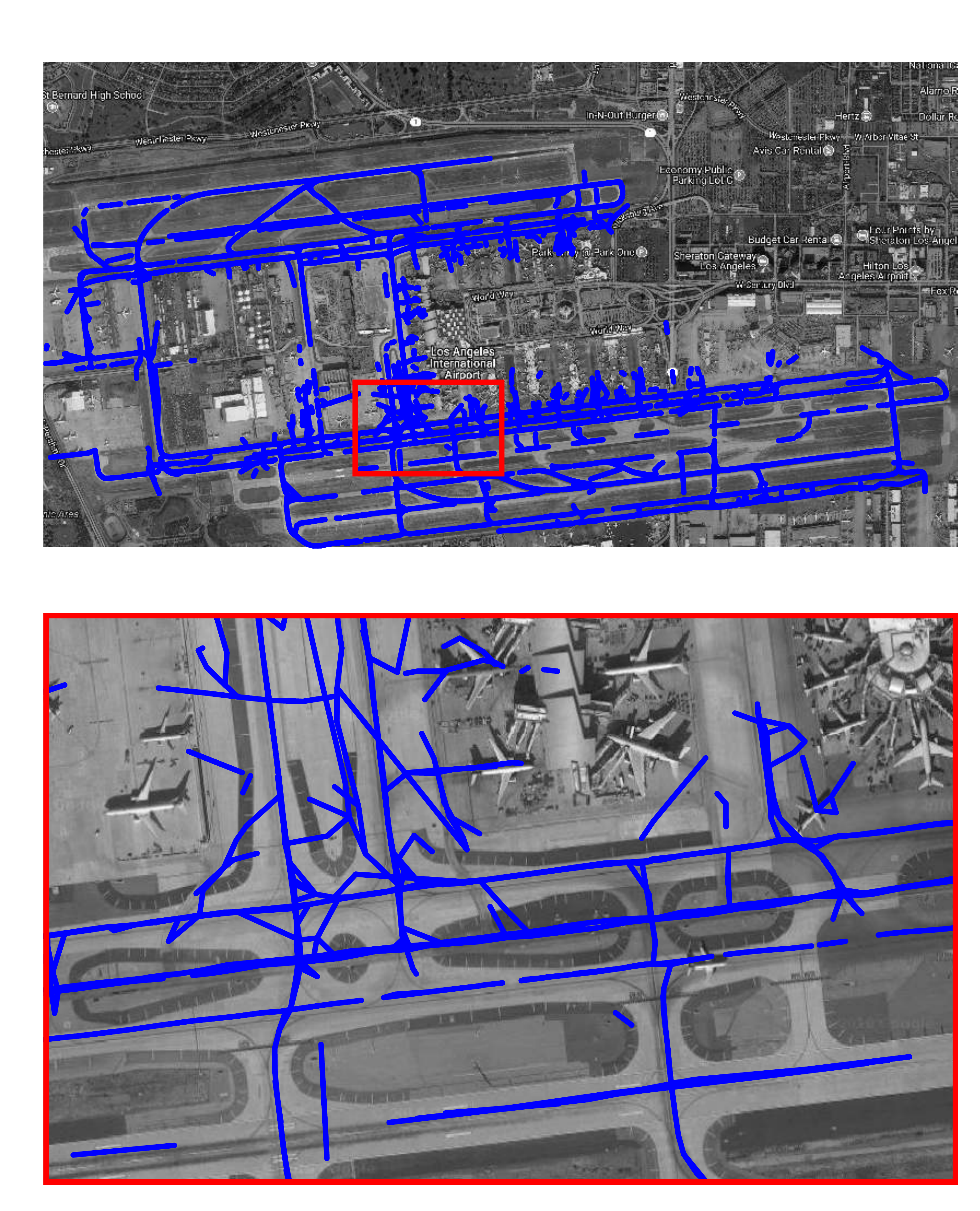}\label{fig:faa_results_a}}
    \subfigure[Chen \etal \cite{chen2016city} on FAA]{
        \includegraphics[width=0.48\columnwidth]{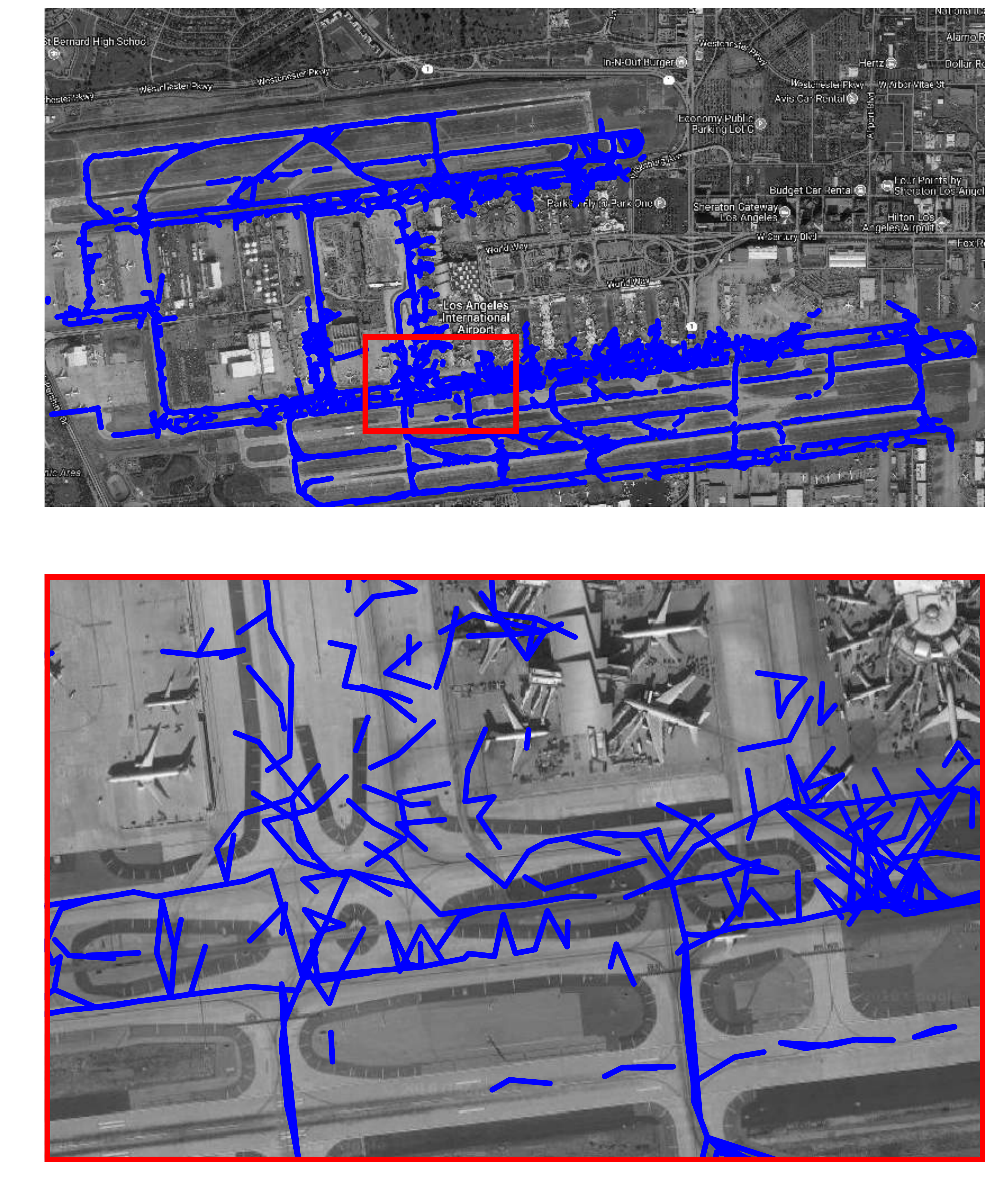}\label{fig:faa_results_c}}
    \caption{Visual evaluations of the inferred map by our COLTRANE and the approach by Chen \etal \cite{chen2016city}. The zoomed area on UIC is a high-built area that exacerbated the GPS noise. The zoomed area on FAA contains a very high junction density. Google map is used as the groundtruth.}
    \label{fig:viz_eval}
\end{figure}


Figure \ref{fig:viz_eval} shows that COLTRANE produces a smoother road map compared to the approach of Chen \etal \cite{chen2016city}. Moreover, COLTRANE produces fewer spurious edges, which is a visible issue in Chen \etal \cite{chen2016city}, particularly for areas with many junctions and urban areas that exacerbated the GPS noise. Our improvements are attributed to the implementation of edge pruning via CNN.

\subsubsection{ITMS}

\begin{figure}[htbp]
\centering
  \subfigure[UIC]{
    \includegraphics[width=0.47\columnwidth]{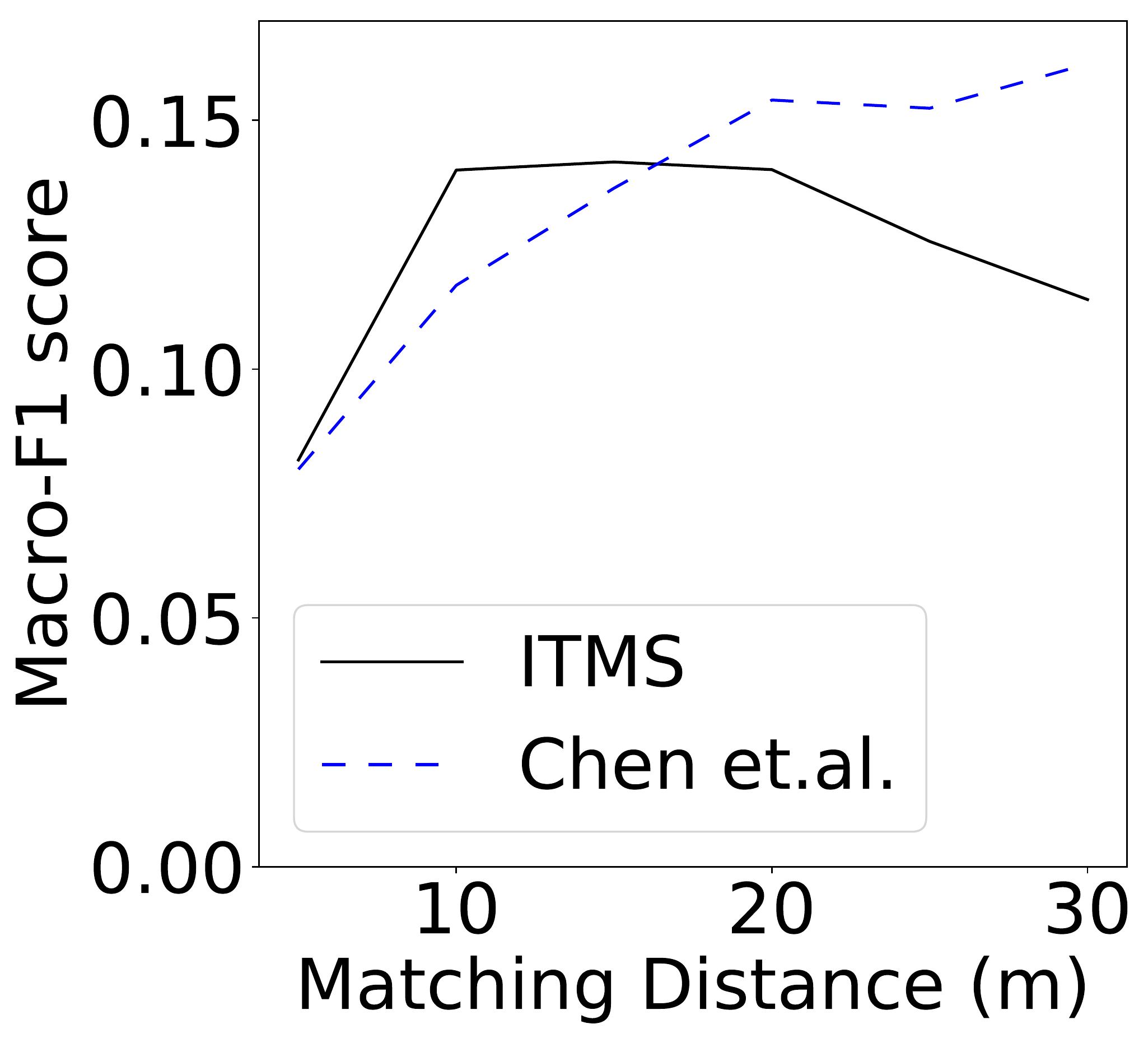}}
  \hspace{0.01\columnwidth}
  \subfigure[FAA]{
    \includegraphics[width=0.47\columnwidth]{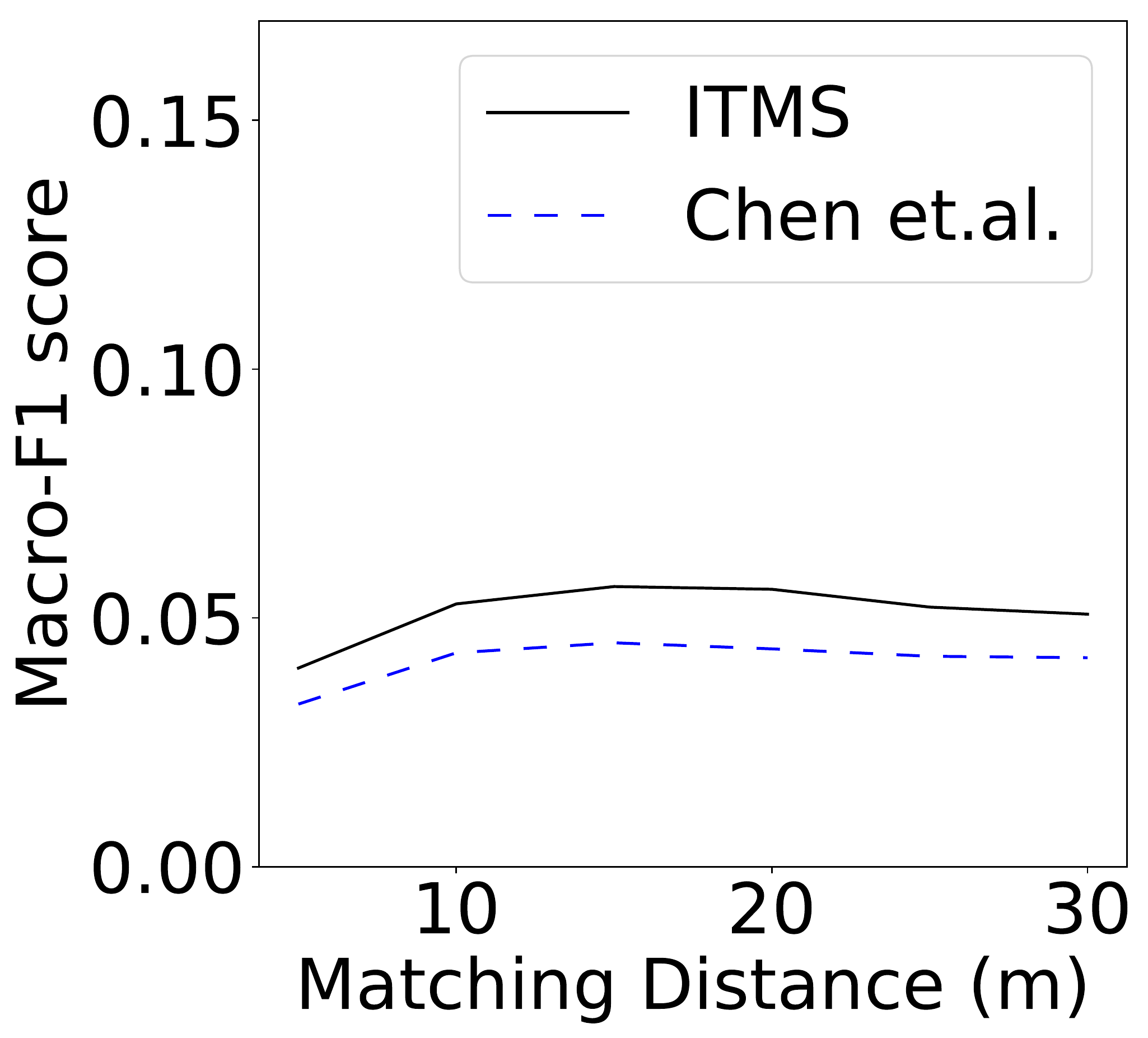}}
  \caption{Performance comparison (Macro--F1) of the map inference between ITMS and the approach of Chen \etal \cite{chen2016city} for a varying matching distance \cite{biagioni2012inferring}.}
  \label{fig:itms_mh}
\end{figure}

Below, we analyze the improvements brought by ITMS when compared to Traj-Meanshift. To do so, we use the algorithm by \cite{chen2016city} and only replace the Trajectory-Meanshift  by ITMS. The result in Figure \ref{fig:itms_mh} shows that although the performance is comparable in the city  environment, ITMS is  performing consistently better in the more spatially complex airport environment.

\subsection{Junction Detection and Classification}

\begin{table}[]
\centering
\caption{Evaluations of junction detection and classification.
}
\label{tab:junction_metrics}
\begin{tabular}{cc|cc}
\toprule
~~\textbf{Dataset}~~ & \textbf{Metric}        & ~~~~~~\textbf{COLTRANE}~~~~~~    & ~~\textbf{Chen \etal \cite{chen2016city}}~~ \\ \midrule \midrule
\multirow{2}{*}{\textbf{UIC}}
     & \textbf{Accuracy}       & \textbf{99.21\%} & 98.24\%          \\ 
    & ~~Macro--F1~~ & \textbf{0.9921}  & 0.9822           \\ \midrule
\multirow{2}{*}{\textbf{FAA}}    & \textbf{Accuracy}       & \textbf{93.73\%} & 92.5\%           \\ 
    & ~Macro--F1~ & \textbf{0.9345}  & 0.9231           \\ \bottomrule
\end{tabular}
\end{table}


Table \ref{tab:junction_metrics} shows similar trend as the previous results. As the airport data is more spatially complex, it yields lower Macro--F1 scores for the case of junction detection and classification. Nevertheless, our proposed COLTRANE framework slightly outperforms the approach of Chen \etal \cite{chen2016city} which is already a strong performer.





\section{Discussion and Limitations}


There exist many ways to improve on COLTRANE. An on-line version of our approach would provide a benefit of the real-time updates. Highlighting dynamic changes could serve a way of anomaly detection to improve the situational awareness; a useful feature for both road users, traffic and aviation authorities.

Moreover, our approach paves an avenue for integration with a more sophisticated spatial data. For instance, we could feed  aerial or satellite images as an input to further improve the accuracy of the inferred map. 
Thus, COLTRANE introduces a novel deep learning approach to processing increasingly ubiquitous trajectory data and other varieties of the spatio-temporal data. 


\section{Conclusions}


We have proposed COLTRANE, a novel deep learning framework for the map inference, junction detection and classification; the first deep learning approach that has been tested in multiple scenarios such as city road network and airport tarmac. We have evaluated our approach on two real-world datasets. COLTRANE has outperformed the approach of \cite{chen2016city} by up to 37\% according to the accuracy and Macro--F1 scores for both the map inference and junction detection/classification tasks. We have also improved upon the road centerline localization algorithm Traj-Mean Shift by proposing ITMS which is more robust to noisy GPS data. Moreover, we have introduced a novel trajectory descriptor for the GPS datapoints  which captures important statistics of GPS datapoints such as occurrences and directional velocities of GPS datapoints. As results show, utilizing CNN to predict the node degree in the graph/map construction yields significant improvements. The degree prediction helps  disambiguate the correct and erroneous predictions of road segments and junctions.


\begin{acks}
This research is partially supported by Northrop Grumman Corporations USA, RMIT University. We would like to also acknowledge the support of the Investigative Analytics team (Data61/CSIRO) and the NVIDIA GPU grant. 
\end{acks}

\bibliographystyle{ACM-Reference-Format}
\bibliography{1mapbib}


\listoftodos{}

\end{document}